\documentclass[final]{cvpr}

\usepackage{times}
\usepackage{color,xcolor}
\usepackage{epsfig}
\usepackage{graphicx}

\usepackage{adjustbox}
\usepackage{array}
\usepackage{booktabs}
\usepackage{colortbl}
\usepackage{float,wrapfig}
\usepackage{hhline}
\usepackage{multirow}
\usepackage{caption}
\usepackage{amsmath,amsfonts,amsthm,amssymb}
\usepackage{bm}
\usepackage{nicefrac}
\usepackage{microtype}

\usepackage{changepage}
\usepackage{extramarks}
\usepackage{fancyhdr}
\usepackage{lastpage}
\usepackage{setspace}
\usepackage{soul}
\usepackage{xspace}
\usepackage{dsfont}

\usepackage{algorithm}
\usepackage{algpseudocode}
\usepackage{enumerate}

\usepackage{url}
\usepackage[pagebackref=true,breaklinks=true,colorlinks,bookmarks=false]{hyperref}
\usepackage{paralist}

\newcommand{\myparagraph}[1]{\vspace{-10pt}\paragraph{#1}}

\newcommand{\ignorethis}[1]{}

\renewcommand*{\thefootnote}{\fnsymbol{footnote}}

\DeclareMathOperator*{\argmin}{arg\,min}

\makeatletter
\DeclareRobustCommand\onedot{\futurelet\@let@token\@onedot}
\def\@onedot{\ifx\@let@token.\else.\null\fi\xspace}

\makeatother

\newcommand{\image}{\mathbf{x}}
\newcommand{\fake}{\Tilde{\image}}

\makeatletter
\newcommand\footnoteref[1]{\protected@xdef\@thefnmark{\ref{#1}}\@footnotemark}
\makeatother

\definecolor{mydarkblue}{rgb}{0,0.08,1}
\definecolor{mydarkred}{rgb}{0.8,0.02,0.02}
\definecolor{mydarkorange}{rgb}{0.40,0.2,0.02}
\definecolor{mypurple}{RGB}{111,0,255}
\definecolor{myred}{rgb}{1.0,0.0,0.0}
\definecolor{mygold}{rgb}{0.75,0.6,0.12}
\definecolor{mydarkgray}{rgb}{0.66, 0.66, 0.66}
\definecolor{mygray}{gray}{0.9}
\definecolor{keynotegreen}{rgb}{0.04,0.52,0}
\definecolor{keynoteyellow}{rgb}{1,0.68,0}

\def\method{Anycost GAN\xspace}

\newcommand{\reffig}[1]{Figure~\ref{fig:#1}}
\newcommand{\refsec}[1]{Section~\ref{sec:#1}}

\newcommand{\lblfig}[1]{\label{fig:#1}}
\newcommand{\lblsec}[1]{\label{sec:#1}}

\newcommand\blfootnote[1]{%
	\begingroup
	\renewcommand\thefootnote{}\footnote{#1}%
	\addtocounter{footnote}{-1}%
	\endgroup
}

\def\cvprPaperID{2535} %
\def\confYear{CVPR 2021}

\begin{document}

\title{Anycost GANs for Interactive Image Synthesis and Editing}

\author{Ji Lin\textsuperscript{1,2}\thanks{}
\qquad
Richard Zhang\textsuperscript{2}
\qquad
Frieder Ganz\textsuperscript{2}
\qquad
Song Han\textsuperscript{1}
\qquad
Jun-Yan Zhu\textsuperscript{2,3}  \\
$^1$MIT \quad $^2$Adobe Research \quad $^3$CMU \\
{\tt\small \{jilin, songhan\}@mit.edu \hspace{0.5pt} \{rizhang, ganz\}@adobe.com \hspace{0.5pt}  \hspace{0.5pt} junyanz@cs.cmu.edu  } \\
}

\twocolumn[{%
\renewcommand\twocolumn[1][]{#1}%
\vspace{-3em}
\maketitle
\vspace{-3em}
\begin{center}
    \centering
    \includegraphics[width=\linewidth]{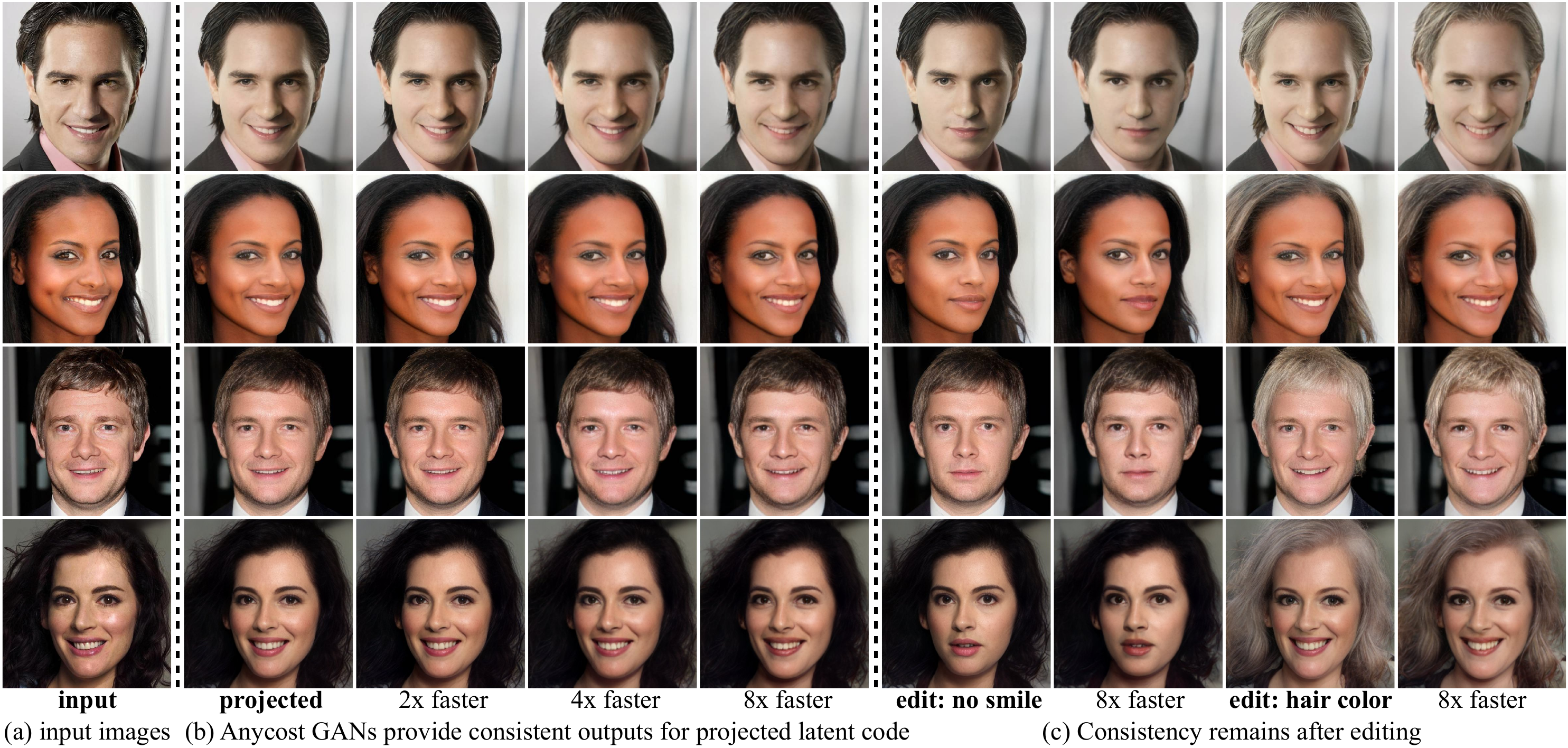}
    \captionof{figure}{
   \method can be executed at flexible computation costs (fast preview with low cost and high-quality output with full cost), enabling interactive image editing with quick preview. The low-cost sub-generator produces consistent outputs compared to the full-cost generator during both image projection and latent code traversal, making the sub-generator an accurate proxy for various editing tasks (\eg, no smile, changing hair color). Interactive demos are available \href{https://youtu.be/_yEziPl9AkM?t=90}{here}. 
    }
    \lblfig{teaser}
\end{center}%
}]

\begin{abstract}
\blfootnote{$*$ Part of the work done during an internship at Adobe Research.}
   Generative adversarial networks (GANs) have enabled photorealistic image synthesis and editing. %
   However, due to the high computational cost of large-scale generators (e.g., StyleGAN2), it usually takes seconds to see the results of a single edit on edge devices, prohibiting interactive user experience.  In this paper, inspired by quick preview features in modern rendering software, we propose \method for interactive natural image editing. 
   We train the \method to support elastic resolutions and channels for faster image generation at versatile speeds. Running subsets of the full generator produce outputs that are perceptually similar to the full generator, making them a good proxy for quick preview.
   By using sampling-based multi-resolution training, adaptive-channel training, and a generator-conditioned discriminator, the anycost generator can be evaluated at various configurations while achieving better image quality compared to separately trained models.
   Furthermore, we develop new encoder training and latent code optimization techniques to encourage consistency between the different sub-generators during image projection.
   \method can be executed at various cost budgets (up to 10$\times$ computation reduction) and adapt to a wide range of hardware and latency requirements.
   When deployed on desktop CPUs and edge devices, our model can provide perceptually similar previews at 6-12$\times$ speedup, 
   enabling interactive image editing. The \href{https://github.com/mit-han-lab/anycost-gan}{code} and \href{https://youtu.be/_yEziPl9AkM}{demo} are publicly available.
\end{abstract}

\vspace{-5mm}

\section{Introduction}

Generative Adversarial Networks (GANs)~\cite{goodfellow2014generative} have excelled at synthesizing diverse and realistic images from randomly sampled latent codes~\cite{karras2018progressive, karras2019style, karras2020analyzing, miyato2018spectral, brock2019large}. Furthermore, a user can transform the generated outputs (\eg, add smiling to a portrait) by tweaking the latent code~\cite{radford2015unsupervised, jahanian2019steerability, karras2019style, harkonen2020ganspace, shen2020interpreting}. In real-world use cases, a user would often like to edit a natural image rather than generating random samples. To achieve this, one can project the image into the image manifold of GANs by finding a latent code that reconstructs the image, and then modify the code to produce final outputs~\cite{zhu2016generative}.

Despite its photorealistic results and versatile editing ability, modern deep generative models incur huge computational costs, prohibiting edge deployment. For example,  the StyleGAN2 generator~\cite{karras2020analyzing}  consumes 144G MACs, 36$\times$ larger compared to ResNet-50~\cite{he2016deep}. The expensive model often introduces a several-second delay for a single edit, leading to a sub-optimal user experience and shorter battery life when used on an edge device.

Modern 2D/3D content creation workflows, such as the preview rendering feature in Maya and Blender, as well as the playback feature in Adobe Premiere Pro, allow users to easily control the tradeoff between image quality and rendering speed. A user can turn off certain visual effects, reduce the resolution and fidelity, or use a fast method during user interaction. Once the edit is finalized, a user can use an expensive method with additional visual effects at a higher resolution. In rendering literature, this tradeoff can be easily achieved by reducing the number of sampled rays in ray/path tracing~\cite{kajiya1986rendering}, or early stopping of iterative solvers in progressive radiosity~\cite{goral1984modeling,cohen1988progressive}. 
 In this work, we aim to bring a smooth tradeoff between visual quality and interactivity to deep generative models.

We propose ``Anycost'' GANs for interactive image synthesis and editing. Our goal is to train a generator that can be executed at a wide range of computational costs while producing visually consistent outputs: we can first use a low-cost generator for fast, responsive previews during image editing, and then use the full-cost generator to render high-quality final outputs. We train the anycost generators to support \emph{multi-resolution} outputs and \emph{adaptive-channel} inference. The smaller generators are nested inside the full generator via weight-sharing. 
Supporting different configurations in one single generator introduces new challenges for minimax optimization, as the adversarial optimization involves too many players (different sub-generators). Vanilla GAN training methods fail catastrophically in this setting. 
To stabilize the training process, we propose to perform stage-wise training: first train sub-generators at multiple resolutions but with full channels, and then train sub-generators with reduced channels. To account for different sub-generators' capacities and architectures, we train a weight-sharing discriminator that is conditioned on the architectural information of the specific sub-generator. 

We train \method to support two types of channel configurations: uniform channel reduction ratio and flexible ratios per layer. The combined architecture space leads to high flexibility in terms of computation cost, containing sub-generators at $>10\times$ computation difference. We can directly obtain low-cost generators by taking a subset of weights without any fine-tuning, which allows us to easily switch between quality and efficiency. 
To handle diverse hardware capacities, we use evolutionary search to automatically find the best sub-generator under different computational budgets, while achieving the best output consistency \wrt the full-cost generator (\reffig{teaser}b). 

To better maintain consistency during the image projection process, we further propose consistency-aware encoder training and iterative optimization for image projection. We optimize the reconstruction loss, not only for the full-cost generator, but also for the sub-generators, which significantly improves the consistency during both image projection and editing steps (\reffig{teaser}c). 

Our single, anycost generator can provide visually consistent outputs at various computation budgets. Compared to small generators of the same architecture or existing compression methods based on distillation, our method can provide better generation quality  and higher consistency \wrt to the full generator.
The generated outputs from generators at different costs also share high-level visual cues, for example, producing generated faces with consistent facial attributes (Table~\ref{tab:attribute}). Our method provides 12.2$\times$ speed-up on Xeon CPU and 8.5$\times$ speed up on Jetson Nano GPU for faster preview.  Combined with consistency-aware image projection, our anycost generator maintains consistency after various editing operations (Figure~\ref{fig:compare_edited}) and offers an efficient and interactive image editing experience. %

\section{Related Work}
\begin{figure*}[t]
    \centering
    \includegraphics[width=0.95\textwidth]{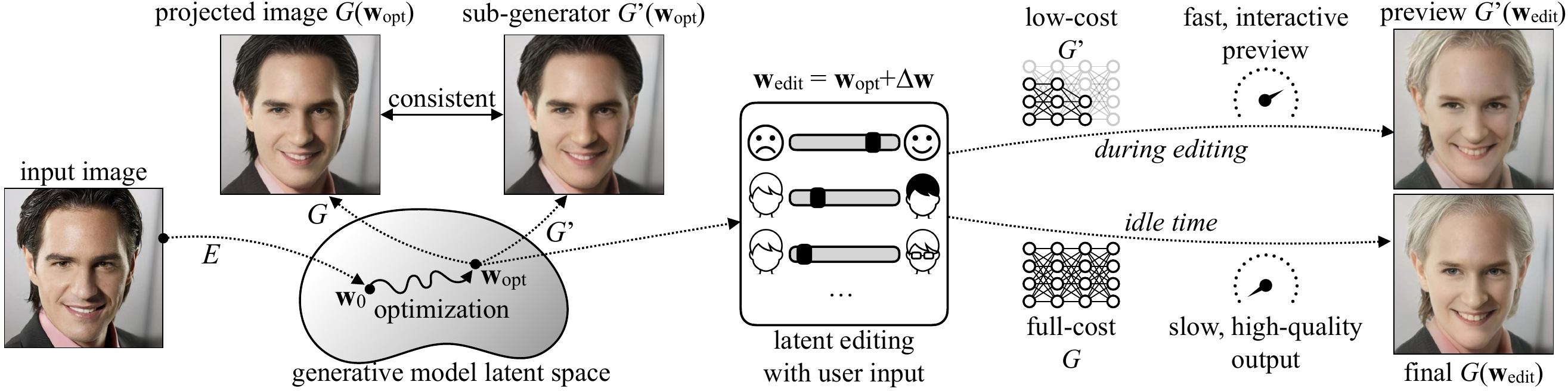}
    \caption{\method for image synthesis and editing. Given an input image, we project it into the latent space with encoder $E$ and backward optimization. We can modify the latent code with user input to edit the image. During editing, a sub-generator of small cost is used for fast and interactive preview; during idle time, the full cost generator renders the final, high-quality output. The outputs from the full and sub-generators are visually consistent during projection and editing. %
      }
    \vspace{-10pt}
    \label{fig:overall}
\end{figure*}

\vspace{4pt}
\myparagraph{Generative Adversarial Networks.} GANs~\cite{goodfellow2014generative} have enabled photorealistic synthesis for many vision and graphics applications~\cite{ledig2017photo,choi2017stargan,isola2017image,zhu2017unpaired,pathak2016context,hoffman2018cycada,ganin2016domain}. Some recent examples include high-resolution face synthesis~\cite{karras2019style,karras2020analyzing}, ImageNet-scale class-conditional generation~\cite{brock2019large}, and semantic photo synthesis~\cite{wang2018pix2pixHD,park2019semantic}. As image quality and model capacity have increased~\cite{goodfellow2014generative,radford2015unsupervised,zhang2017stackgan,karras2018progressive,zhang2018self,karras2020analyzing,Karnewar_2020_CVPR}, so have computational costs and inference time. For example, it takes several seconds for a single forward pass of StyleGAN2~\cite{karras2020analyzing} on a desktop CPU or a tablet. While most research demos are running on the latest GPUs, deploying them on edge devices and laptops efficiently is critical for interactive editing applications. In this work, we tackle this problem by learning hardware-adaptive generators that work across a wide range of devices and latency constraints.

\myparagraph{Model acceleration and dynamic networks.} Efficient deep networks have enabled fast inference and reduced model sizes with a focus on image classifiers~\cite{han2015learning,han2015deep,zhu2016trained,wang2019haq,han2019design}. Commonly used approaches include pruning~\cite{he2017channel,lin2017runtime,liu2017learning, han2015learning,han2015deep,wen2016learning}, quantization~\cite{han2015deep,zhu2016trained}, knowledge distillation~\cite{hinton2015distilling,luo2016face,chen2017learning}, efficient architecture design~\cite{chollet2017xception,iandola2016squeezenet,howard2017mobilenets,sandler2018mobilenetv2}, and autoML-based methods~\cite{he2018amc,howard2019searching, wang2019haq, zoph2016neural, liu2018darts, cai2018proxylessnas,guo2020single, lin2020mcunet}. Recently, several works adopt the above ideas to compress generative networks with a focus on image-conditional GANs~\cite{shu2019co,li2020gan,yu2020self,wang2020gan,fu2020autogan}, such as pix2pix~\cite{isola2017image} and CycleGAN~\cite{zhu2017unpaired}. The work most related to ours is Aguinaldo et al.~\cite{aguinaldo2019compressing}, which adopts knowledge distillation to compress DCGAN~\cite{radford2015unsupervised} on low-res images of MNIST, CIFAR, and CelebA datasets. A concurrent work~\cite{hou2020slimmable} also explores channel reduction of unconditional GANs on low-resolution datasets. Our experiments show that knowledge distillation alone is insufficient to transfer the knowledge of a GAN on large-scale datasets (Figure~\ref{fig:fid_vs_macs}). Compared to prior work, our method works well for large-scale, high-resolution unconditional GANs. More importantly,  our method can produce output images with dynamic resolution and fidelity, not possible by the prior work~\cite{aguinaldo2019compressing}. We also include techniques to leverage the anycost generator in real image editing scenarios.

\myparagraph{Image editing via GANs.} There exist two major ways of using GANs for image editing: (1) conditional GANs~\cite{isola2017image,zhu2017unpaired,liu2017unsupervised,liu2020generative,lee2018diverse}, which learn to directly translate an input image into a target domain, and (2) image projection~\cite{zhu2016generative,brock2017neural}, where the algorithm first projects a real image into the latent space of an unconditional GAN, modifies the latent code to achieve an edit, and synthesizes a new image accordingly. Recently, interest in projection-based editing has been revived due to the increasing quality of unconditional GANs. Several methods have been proposed, including choosing better or multiple layers to project and edit~\cite{abdal2019image2stylegan,abdal2020image2stylegan++,gu2020image}, fine-tuning network weights for each image~\cite{bau2019semantic}, modeling image corruption and transformations~\cite{anirudh2020mimicgan,huh2020transforming}, and discovering meaningful latent directions~\cite{shen2020interpreting,goetschalckx2019ganalyze,jahanian2019steerability,harkonen2020ganspace}.
In this work, we aim to learn efficient generators for downstream image projection and editing, which offers unique challenges. Namely, we wish the projected latent code from a sub-generator provide a faithful ``preview'' of the full model. In addition, user edits with the sub-generator should also make the corresponding changes to the full model. We tackle these challenges using our new consistency-aware encoder training and optimization method.

\section{Anycost GANs}

We introduce the problem setting and potential challenges in \refsec{problem}. We then describe our training method for learning multi-resolution adaptive-channel generators in \refsec{learning}. In \refsec{projection}, we describe our consistency-aware encoder training and latent code optimization, designed for image projection and editing.

\subsection{Problem Setup}
\lblsec{problem}

An unconditional generator $G$ learns to synthesize an image $\image \in \mathds{R}^{H\times W \times 3}$ given a random noise %
 $\mathbf{z} \in \mathcal{Z}$, where $\mathcal{Z}$ is a low-dimensional latent space. In this work, we study StyleGAN2~\cite{karras2020analyzing}, which first learns a non-linear mapping $f: \mathcal{Z}\rightarrow\mathcal{W}+$ that produces $\mathbf{w}=f(\mathbf{z}), \mathbf{w}\in \mathcal{R}^{18\times 512}$, and then generates the image from 
 $\mathcal{W}+$
 space: $\image=G(\mathbf{w})$~\cite{abdal2019image2stylegan}.

Anycost GAN enables a subsection of the generator $G'$ to execute independently and produce a similar output $\image'=G'(\mathbf{w})$ to the full generator $G(\mathbf{w})$. We can then use $G'(\mathbf{w})$ for fast preview during interactive editing, and full $G(\mathbf{w})$ to render final high-quality outputs.
The model can be deployed on diverse hardware (\eg, GPUs, CPUs, smartphones), and users can choose between different preview qualities.

A natural choice for enabling diverse inference cost is to use different image resolutions, as done in StackGAN~\cite{han2017stackgan} and ProGAN~\cite{karras2018progressive}. However, reducing resolution from 1024$\times$1024 to 256$\times$256 only reduces the computation by $1.7\times$, despite $16\times$ fewer pixels to synthesize\footnote{1024-resolution StyleGAN2 has 144G MACs, while 256-resolution model has 85G MACs.}. Therefore, we need additional dimensions to further reduce the cost.

\subsection{Learning Anycost Generators}
\lblsec{learning}

\begin{figure}[t]
    \centering
    \includegraphics[width=1\linewidth]{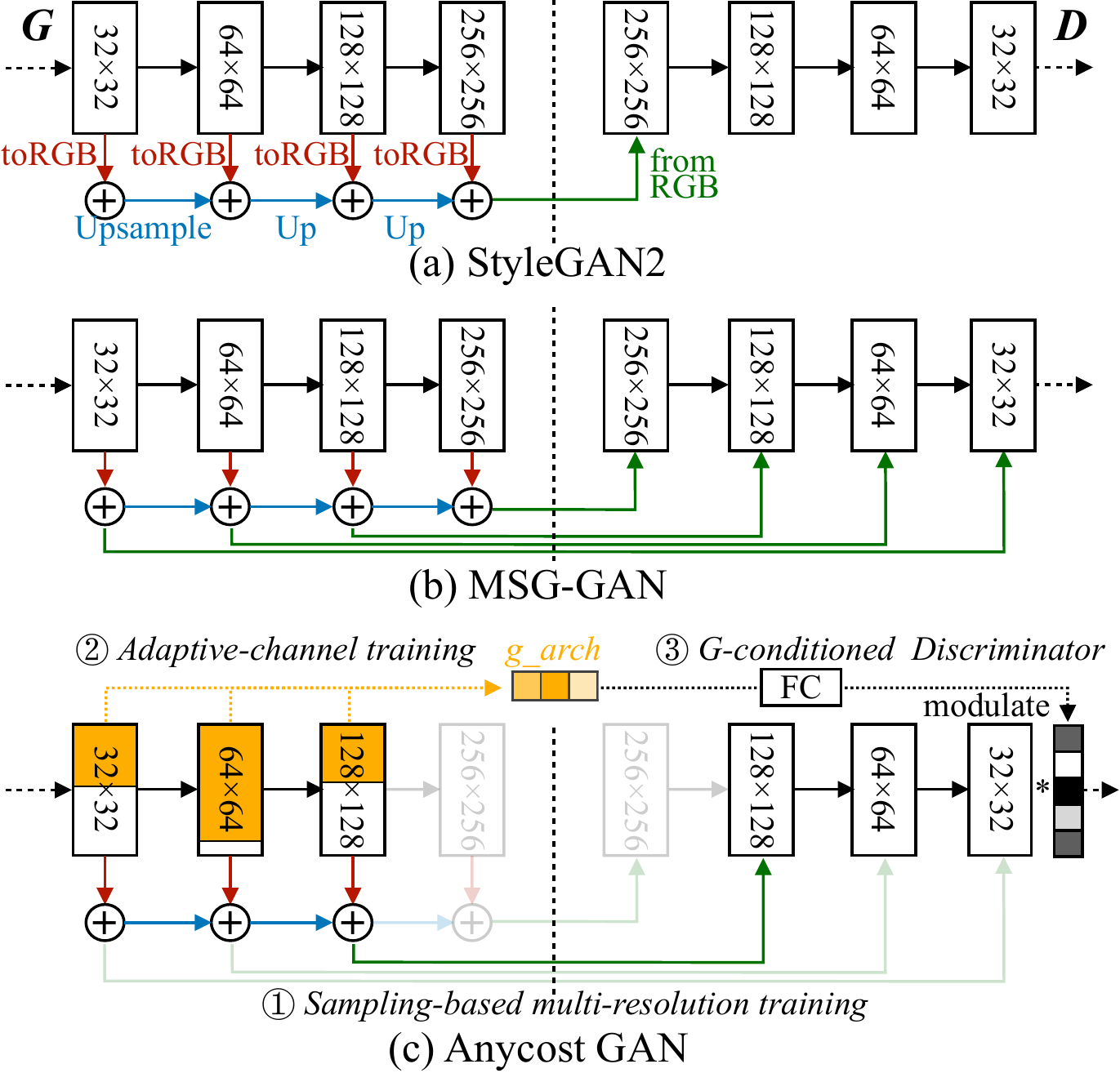}
    \caption{Anycost GAN synthesizes realistic images at a  wide range of resolutions and model capacities via (1) sampling-based multi-res training, (2) adaptive-channel training, and (3) generator-conditioned discriminator.
      }
    \vspace{-10pt}
    \label{fig:detailed_arch}
\end{figure}

We propose to learn an anycost generator to produce proxy outputs at diverse \emph{resolutions} and \emph{channel capacities}. 
The detailed architecture is shown in Figure~\ref{fig:detailed_arch}.

\myparagraph{Multi-resolution training.}
For elastic resolutions, the architectures of the ProGAN~\cite{karras2018progressive} and StyleGAN family~\cite{karras2019style,karras2020analyzing} already produce lower-res intermediate outputs, although the outputs do not look natural (\reffig{multires_distill}a). We can enable lower-resolution outputs by enforcing multi-scale training objectives (\reffig{multires_distill}b).
Our generator gradually produces higher resolution outputs after each block $g^k$:
\begin{equation}
    \fake = G(\textbf{w}) = g^K \circ g^{K-1} \circ ... \circ g^{k}\circ ...\circ g^2 \circ g^1(\textbf{w}),
\end{equation}
where $K$ is the total number of network blocks.
We use the intermediate low-res outputs for preview, and denote:
\begin{equation}
\fake^k = G^k(\textbf{w}) = g^k\circ g^{k-1} \circ ... \circ g^2 \circ g^1(\textbf{w}), k\leq K, 
\end{equation}
resulting a series of outputs in increasing resolutions $[\fake^1, ..., \fake^K]$.

Existing work MSG-GAN~\cite{Karnewar_2020_CVPR} trains the generator to support different resolutions by using a discriminator taking images of all resolutions at the same time (Figure~\ref{fig:detailed_arch}b). However, such an all-resolution training mechanism leads to lower quality results on large-scale datasets like FFHQ~\cite{karras2020analyzing} (as measured by Fr\'echet Inception Distance, FID~\cite{heusel2017gans}), compared to training single-resolution models separately (Table~\ref{tab:multires_fid}). To overcome the image fidelity degradation when supporting multi-resolutions, we propose a \emph{sampling-based} training objective, where a \emph{single} resolution is sampled and trained at each iteration, both for the generator $G$ and the discriminator $D$. 
As shown in Figure~\ref{fig:detailed_arch}c, when sampling a lower resolution (\eg, 128$\times$128), the translucent parts are not executed.
We use the intermediate output of $G$ for a lower resolution. It is passed through a \textcolor{keynotegreen}{fromRGB} convolution ``readoff'' layer to increase channels, and then fed to an intermediate layer of $D$. 
Our multi-resolution training objective is formulated as:
\begin{equation}
    \mathcal{L}_{\text{multi-res}} = \mathbb{E}_{\image, k}[\log D(\image^k)] + \mathbb{E}_{\mathbf{w}, k}[\log(1-D(G^k(\textbf{w})))].
\end{equation}
We observe that the \emph{sampling-based} training leads to better convergence: in fact, our multi-resolution model gives better FID at all resolutions, compared to single-resolution models trained at corresponding resolutions, as shown in Table~\ref{tab:multires_fid}. %

\begin{figure}[t]
    \centering
    \includegraphics[width=1\linewidth]{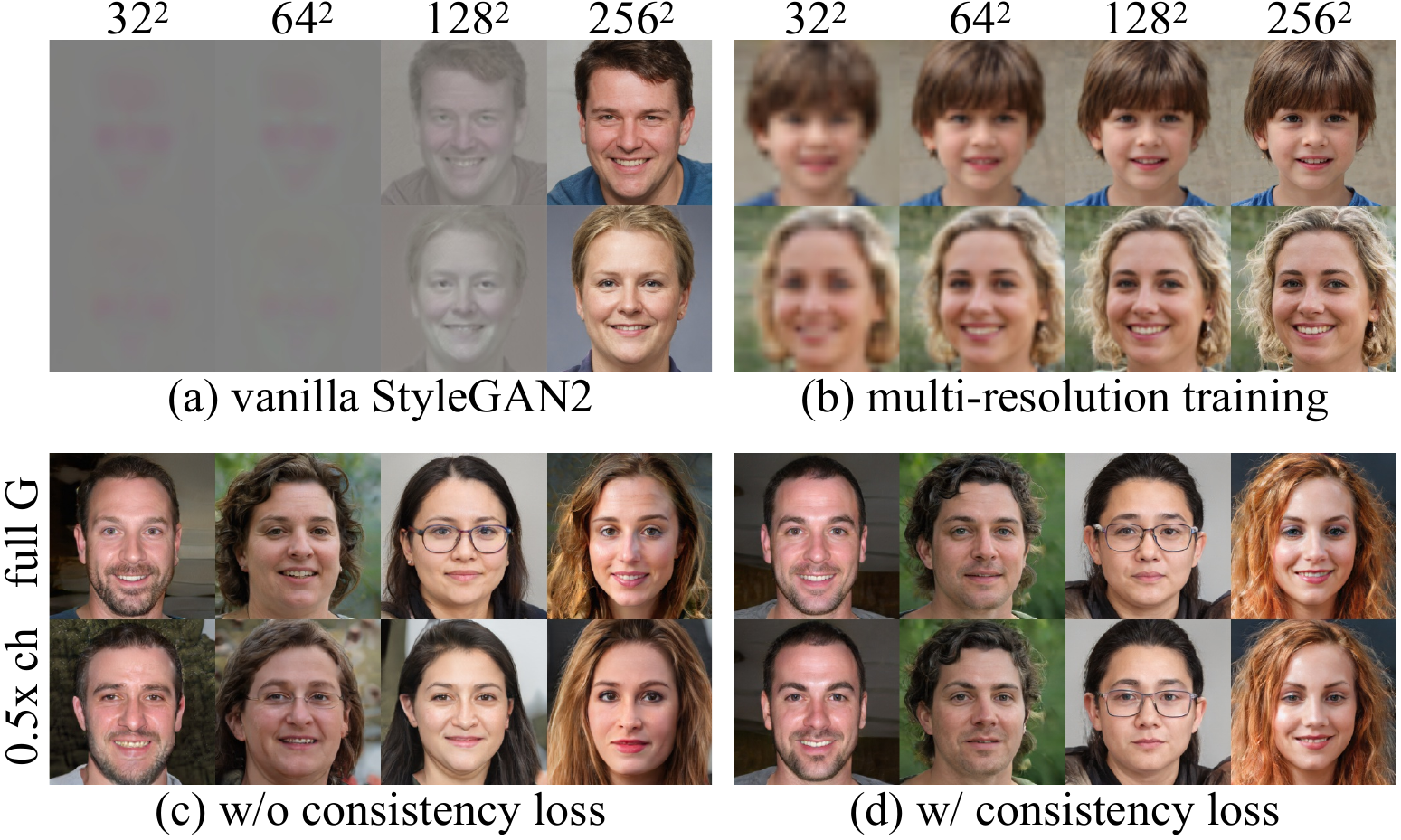}
    \caption{(a) Intermediate layers do not output realistic images for StyleGAN2 baseline. (b) Multi-resolution training in our model enables photorealistic lower-resolution outputs at intermediate layers. (c) Without consistency loss, running the model at full and half-channel produces inconsistent images. (d) Adding consistency loss maintains output consistency (similar images per column).
      }
    \lblfig{multires_distill}
\end{figure}

\myparagraph{Adaptive-channel training.}
To further enable our generator to run at different costs, we train the generator to support variable channels.
For adaptive-channel training, we allow different channel number multipliers for each layer (either a \emph{uniform} ratio, used across all layers or \emph{flexible} ratios for each layer). For each training iteration, we randomly sample a channel multiplier configuration and update the corresponding subset of weights (\textcolor{keynoteyellow}{yellow} part in Figure~\ref{fig:detailed_arch}c).
We hope that the most ``important'' channels are preserved during sampling to minimize any degradation. To this end, we initialize our model using the multi-resolution generator from the previous stage and sort the channels of convolutional layers according to the magnitude of kernels, from highest to lowest.
We always sample the most important $\alpha c$ ones according to the initial sorting, where $\alpha\in [0.25, 0.5, 0.75, 1]$ and $c$ is the number of channels in the layer.
The adaptive-channel training objective is written as follows:
\begin{equation}
    \mathcal{L}_{\text{ada-ch}} = \mathbb{E}_{\image, k}[\log D(\image^k)] + \mathbb{E}_{\mathbf{w}, k, \mathbb{C}}[\log(1-D(G^k_\mathbb{C}(\textbf{w})))],
\end{equation}
where $\mathbb{C}$ means the channel configurations for each layer. 

With such a learning objective, all the sub-networks with fewer channels can generate reasonable images. However, these images may be visually different compared to the full network, failing to provide an accurate ``preview'' (Figure~\ref{fig:multires_distill}c). To keep the output consistent across different sub-networks, we add the following consistency loss:
\begin{equation}
    \mathcal{L}_{\text{total}} = \mathcal{L}_{\text{ada-ch}} + \mathbb{E}_{\textbf{w}, k, \mathbb{C}}[\ell (G^k_{\mathbb{C}}(\mathbf{w}),  G(\mathbf{w}))],
\end{equation}
where $\ell$ is a pre-defined distance metric. We use a combination of MSE loss and LPIPS loss~\cite{zhang2018unreasonable}, which empirically gives the most visually consistent results (\reffig{multires_distill}d). Note that all the sub-networks and the full network share weights across different channels and are jointly trained. %

\myparagraph{Generator-conditioned discriminator.}

Unlike conventional GAN training, our system trains many sub-generators of different channels and resolutions at the same time. We find that one single discriminator cannot provide good supervision for all sub-generators of different channel configurations, resulting in an inferior image fidelity/FID (Table~\ref{tab:conditioned_fid}). 
To handle the challenge, we design the discriminator to be \emph{conditioned on the generator architecture}.

A straightforward method for generator-conditioning is to correspondingly shrink the channels of the discriminator with the generator. However, we find that such practice often sacrifices the quality of smaller channel settings, due to its reduced capacity of the discriminator (``reduced ch'' in Table~\ref{tab:conditioned_fid}). Also, such a technique only supports uniform channel ratios.
Instead, we take a learning-based approach to implement the conditioning. 
As shown in Figure~\ref{fig:detailed_arch}c, we first encode the channel configuration in $g\_arch$ using a one-hot encoding (for each layer, we can choose one of the four ratios, forming a one-hot vector of length 4; we concatenate the vectors from all layers to form $g\_arch$), which is passed through a fully connected layer to form the per-channel modulation. The feature map is modulated using the conditioned weight and bias before passing to the next layer. For real images, we randomly draw a $g\_arch$ vector. To stabilize training, we only apply the G-conditioned modulation units to the last two blocks of the discriminator.

\myparagraph{Searching under different budgets.}
By training the generator to support flexible ratios for each layer, we support an exponential number of sub-generator architectures.
At a given computation budget, selecting the proper sub-generator configuration is important for keeping generation quality and consistency.
We use an evolutionary search algorithm~\cite{real2019regularized, cai2020once, guo2020single} to find an effective sub-generator architecture under diverse resource budgets (\eg, computation, latency). Given a certain budget, our evolutionary search minimizes the difference between the desired sub-generator and the full generator's outputs, measured by a perceptual loss. The details can be found in Section~\ref{sec: supp_evolve}.

\subsection{Image Projection with Anycost Generators}
\lblsec{projection}

To edit an existing image $\image$, we need to first project the image into the latent space of a generator~\cite{zhu2016generative} by solving  $\mathbf{w}^*=\argmin_{\mathbf{w}} \ell{( G(\mathbf{w}),  \image})$, where we use a combination of LPIPS~\cite{zhang2018unreasonable} and MSE loss for $\ell$.
We follow \cite{abdal2019image2stylegan,abdal2020image2stylegan++} to project the image into the extended $\mathcal{W}+$ space, rather than the $\mathcal{Z}$ space due to its better expressiveness. We follow the two-step approach, as introduced in iGAN~\cite{zhu2016generative}: encoder-based initialization followed by gradient-based latent code optimization.

\myparagraph{Consistency-aware image projection.}
\emph{Encoder-based} projection directly trains an encoder $E$ for projection by optimizing $E^*=\arg \min_{E} \mathbb{E}_{\image}\ell (G(E(\image)), \image)$ over many training images. For a specific image sample, we can further improve the results with \emph{optimization-based} projection by solving $\textbf{w}^* = \argmin_{\mathbf{w}}\ell (G(\mathbf{w}), \image)$ with iterative gradient descent. 
While our generator can produce consistent results across sub-generators for a \textit{randomly} sampled latent code, the predicted/optimized latent codes $E(\mathbf{x})$ may not follow the prior distribution. As a result, the sub-generators may not produce consistent results on some optimized latent codes. 
Therefore, we modify the objects to produce a latent code that works for both the full generators as well as randomly sampled sub-generators as follows:
\begin{align}
 E^*&=\arg \min_{E} \mathbb{E}_{\image}[\ell (G(E(\image)), \image) + \alpha \mathbb{E}_{k, \mathbb{C}} \ell (G^k_{\mathbb{C}}(E(\image)), \image)] \\
 \textbf{w}^* &= \argmin_{\mathbf{w}}[\ell (G(\mathbf{w}), \image) + \alpha \mathbb{E}_{k, \mathbb{C}} \ell (G^k_{\mathbb{C}}(\mathbf{w}, \image)]
\end{align}
We set hyper-parameter $\alpha=1$ in our experiments.

\myparagraph{Image editing with anycost generators.}
After projection, we can perform image editing by simply changing the latent code and synthesize a new one using $G(\mathbf{w}+\Delta \mathbf{w})$, where $\Delta \mathbf{w}$ is a vector that encodes a certain change. Several methods ~\cite{shen2020interpreting,harkonen2020ganspace} have been proposed to discover such latent directions that control certain aspects of the input (\eg, smiling/non-smiling for faces,  color, and shape for cars). To produce a preview with low latency, we can run $G^k_{\mathbb{C}}(\mathbf{w}+\Delta \mathbf{w})$. In experiments, we observe that as long as the initial projection is consistent across the full and sub-generators, the edited results are visually similar. 
\section{Experiments}

\subsection{Setup}
\vspace{6pt}

We conduct experiments on both FFHQ~\cite{karras2019style} (resolution 1024) and LSUN car dataset~\cite{yu2015lsun} (resolution 512) due to the large scale and high resolution.
Our generators are based on StyleGAN2~\cite{karras2020analyzing} (config-F).  
We train our models to support four resolutions from the highest (\eg, 1024, 512, 256, 128 for FFHQ), since lower resolutions are too blurry for a high-quality preview.
For dynamic channels,
we support multipliers ranging
[0.25, 0.5, 0.75, 1]
\wrt the original channel numbers. For fast ablation studies, we train models on 256$\times$256 images using config-E (half channels). 
More experimental details are provided in Section~\ref{sec: supp_exp_details}.

For adaptive channel support, we consider two settings: (1) training with uniform channel ratios (\ie, using the same channel reduction ratios for all layers); (2) training with flexible channel ratios for each layer. For the latter setting, after GAN training, we leverage evolutionary search to find the best channel \& resolution configurations under certain computation budgets. The details of the evolutionary search can be found in Section~\ref{sec: supp_evolve}.

\subsection{Ablation studies}

We first show how we improve the image quality while supporting different resolutions and channels with ablation studies. The results are evaluated on FFHQ dataset.

\myparagraph{Multi-resolution training.}
\begin{table}[t]
    \setlength{\tabcolsep}{6pt}
    \caption{FID-70k on FFHQ of different multi-resolution training techniques. Our sampling-based technique can train one model that produces multiple resolution outputs with higher image quality (measured by FID~\cite{heusel2017gans}) compared to single resolution training. The models are trained with half channels (Config-E) for faster ablation.}
    \label{tab:multires_fid}
    \centering
    \small{
     \begin{tabular}{lcccccc}
    \toprule
      Resolution & 1024 & 512 & 256 & 128 & 64 & 32  \\  
    \midrule
    Single-resolution  & 3.25  & 4.17 & 3.76 & 4.04 & 3.32 & 2.41 \\ 
    MSG-GAN~\cite{Karnewar_2020_CVPR} & - & - & 4.79 & 6.34 & 2.7 & 3.04 \\
    \midrule
    Ours (low res) & - & - & 3.49 & \textbf{3.26} & \textbf{2.52} & \textbf{2.18} \\
     Ours (high res) & \textbf{2.99} & \textbf{3.08} & \textbf{3.35} & 3.98  & - & -\\
    \bottomrule
     \end{tabular}
     }
\end{table}
We compare the results of sampling-based multi-resolution training against single-resolution training and MSG-GAN~\cite{Karnewar_2020_CVPR} in Table~\ref{tab:multires_fid}. With our technique, we can train one generator that generates multi-resolution outputs at a better quality (lower FID) for both high-resolution (1024) and lower-resolution (256) settings, even compared to specialized single-resolution models. Compared to MSG-GAN~\cite{Karnewar_2020_CVPR}, which trains all resolutions at each iteration, our method consistently gains better FIDs. We hypothesize that feeding images of all resolutions to the discriminator poses a stricter requirement for the generator (all the outputs have to be realistic to fool the discriminator), breaking the balance between the generator and the discriminator.

\myparagraph{G-conditioned discriminators for dynamic channels.}

\begin{table}[t]
    \setlength{\tabcolsep}{2pt}
    \caption{FIDs on FFHQ at different resolutions and channels. All the settings except ``vanilla'' only train one generator and evaluate it at multiple configurations. We mark a better FID \textcolor{keynotegreen}{green} and a worse FID \textcolor{red}{red}  compared to multi-G baseline. Conditioned discriminator (``conditioned'') provides the best FID over different channel widths and resolutions. The model is based on Config-E for faster ablation.}
    \vspace{-5pt}
    \label{tab:conditioned_fid}
    \centering
    \small{ \scalebox{0.92}{
     \begin{tabular}{lcccccccc}
    \toprule
    \multirow{2}{*}{FID-70k$\downarrow$} & \multicolumn{4}{c}{resolution 256} & \multicolumn{4}{c}{resolution 128} \\ \cmidrule(lr){2-5} \cmidrule(lr){6-9}
       & 1$\times$ & 0.75$\times$ & 0.5$\times$ & 0.25$\times$ & 1.0$\times$ & 0.75$\times$ & 0.5$\times$ & 0.25$\times$  \\  
    \midrule
    vanilla & 3.80 & 4.64 & 6.20 & 10.39 & 4.04 & 4.99 & 5.78 &  11.15\\
    \midrule
    same D & \textcolor{keynotegreen}{3.63} & \textcolor{keynotegreen}{3.91} & \textcolor{keynotegreen}{5.41} & \textcolor{red}{14.01} & \textcolor{red}{7.25}  & \textcolor{red}{6.81} & \textcolor{red}{5.92} & \textcolor{keynotegreen}{7.57} \\
     reduced ch & \textcolor{keynotegreen}{3.67} &  \textcolor{keynotegreen}{4.35} & \textcolor{keynotegreen}{5.82} & \textcolor{red}{10.62} & \textcolor{keynotegreen}{3.09} & \textcolor{keynotegreen}{3.65} & \textcolor{keynotegreen}{4.74} & \textcolor{keynotegreen}{8.82} \\
    conditioned (Ours) & \textcolor{keynotegreen}{3.73} & \textcolor{keynotegreen}{3.86} & \textcolor{keynotegreen}{4.64} & \textcolor{keynotegreen}{8.06} & \textcolor{keynotegreen}{3.30} & \textcolor{keynotegreen}{3.28}  &  \textcolor{keynotegreen}{3.69}  & \textcolor{keynotegreen}{5.55} \\
    \bottomrule
     \end{tabular} }
     }
     \vspace{-5pt}
\end{table}
Using one fixed discriminator is not enough to handle a set of generators at different channel capacities and resolutions. 
We use a simpler setting for ablation by only supporting four uniform channel reduction ratios, \ie, using the same channel multiplier for all layers. 
As shown in Table~\ref{tab:conditioned_fid}, using the same discriminator for all sub-generators (denoted as ``same D'') cannot consistently match the FIDs compared to single generators trained for a specific resolution and channel width (denoted as ``vanilla''). It also leads to an unstable FID distribution (\eg, for resolution 128, 1.0$\times$ channel gives worse FID compared to 0.5$\times$, despite increased computational resources), since a single discriminator can only provide suitable supervisions for a small subset of sub-generators. 
To improve the performance of each generator, we implement the discriminator to be 
\emph{conditioned} on the generator architecture. A straight-forward method is to also reduce the channels of the discriminator using the same ratio as the generator (denoted as ``reduced ch'' in Table~\ref{tab:conditioned_fid}). This improves the FIDs under some conditions and makes the FIDs monotonic as a function of computation
(wider sub-generators give better FID). However, the narrow generators (\eg, 0.25$\times$) have degraded FIDs due to limited discriminator capacity. This also does not work for non-uniform channel ratio settings, where each layer can use a different channel ratio, leading to exponential combinations.
Instead, our conditioned discriminator (denoted as ``conditioned'') uses a \emph{learned} modulation according to the generator architecture (represented as an architecture vector), without reducing the discriminator capacity. For different channels and resolutions, it provides consistently better FIDs compared to the multi-generator baseline, and also improves over the reduced-channel method.

\myparagraph{Outperforming compression baselines.} 
\begin{figure*}[t]
    \centering
    \includegraphics[width=1\textwidth]{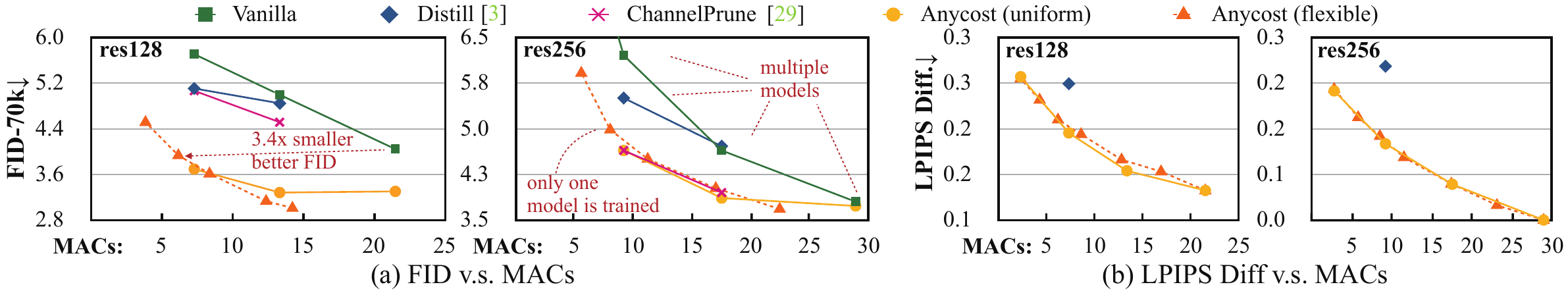}
    \caption{Anycost GAN outperforms existing compression baselines~\cite{aguinaldo2019compressing, he2017channel} at various computation budgets, despite only training a single, flexible generator across computation budgets. }
    \label{fig:fid_vs_macs}
\end{figure*}

One baseline for fast generation preview is to train a small model that mimics the full generator and use the small model for interactive image editing. However, this method does not allow flexible model capacity to fit diverse hardware platforms and latency requirements. The small models also have inferior generation quality and consistency \wrt to the full generator, compared to our anycost generator.

We compare the FID-computation trade-off with compression-based baselines in Figure~\ref{fig:fid_vs_macs}a. ``Vanilla'' means single models trained from scratch.
We compare with~\cite{aguinaldo2019compressing}, which uses a distillation-based method for \emph{unconditional} GAN compression (``Distill''), and also adapt a general CNN compression method~\cite{he2017channel} to GAN (``ChannelPrune''). 
Since the FID cannot be computed across different resolutions, we provide the results for resolution 256 and 128 separately (we can search over all resolutions and channels when optimizing for consistency, like in Figure~\ref{fig:qualititive_flexible}).
For anycost GAN, we provide the results of two settings: uniform channel ratio (denoted as ``Anycost (uniform)'') and flexible channel ratios (denoted as ``Anycost (flexible)''). Anycost GAN outperforms existing compression methods at a \emph{wide range} of computation budgets, despite using only a \emph{single} generator, achieving lower FID/difference at the same computation. The ``uniform'' and ``evolve'' settings achieve a similar trade-off curve. However, with evolutionary search, we can support more fine-grained computation budgets to fit more deployment scenarios.

Apart from generation quality, our sub-generators also provide more consistent outputs \wrt to the full generator (Figure~\ref{fig:fid_vs_macs}b). We measure the LPIPS difference~\cite{zhang2018unreasonable} between the generated images from sub-networks and the full network. Compared to distillation-based training, our model reduces the LPIPS difference by half. For models trained from scratch, we cannot report LPIPS consistency, as the models are independently trained. %

\subsection{Anycost Image Synthesis}
\vspace{6pt}

In this section, we provide the results of Anycost GAN trained on high-resolution images ($1024\times1024$ for FFHQ~\cite{karras2020analyzing} and $512\times384$ for LSUN Car~\cite{yu2015lsun}). The models are trained with Config-F~\cite{karras2020analyzing} for high-quality synthesis.

\myparagraph{Qualitative results.}
\begin{figure}[t]
    \centering
    \includegraphics[width=0.48\textwidth]{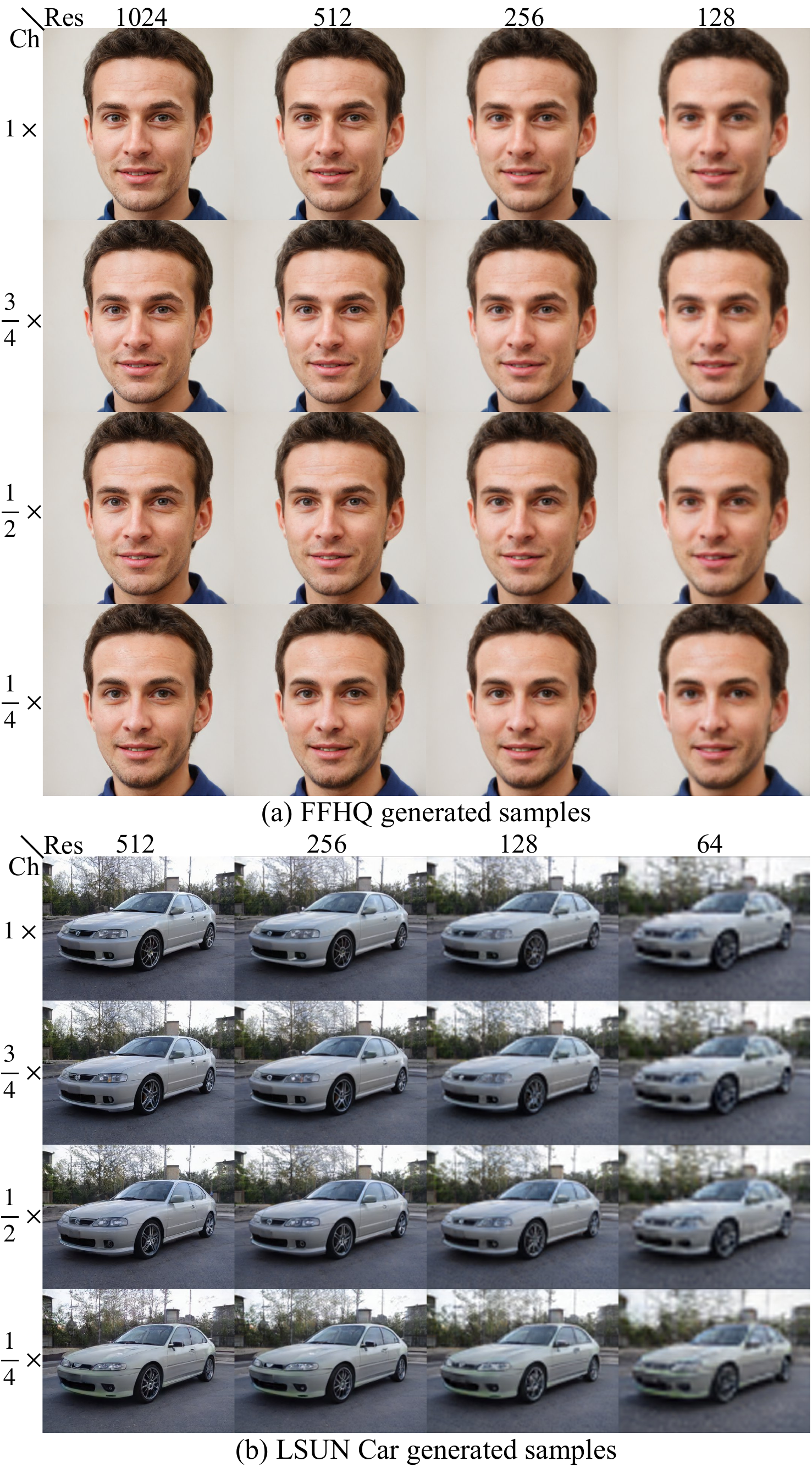}
    \vspace{-15pt}
    \caption{
    \textbf{\method (uniform)} maintains output consistency across different resolutions and channels widths. Zoom in for better view.  For faces (a), our method preserves the prominent facial structure such as age, hair style, and pose. Some small details such as wrinkles are omitted. For cars (b), our method succeeds to keep the color, shape, and pose. Some small details are omitted such as license plates. 
    }
    \vspace{-10pt}
    \label{fig:qualititive}
\end{figure}

\begin{figure}[t]
    \centering
    \includegraphics[width=0.5\textwidth]{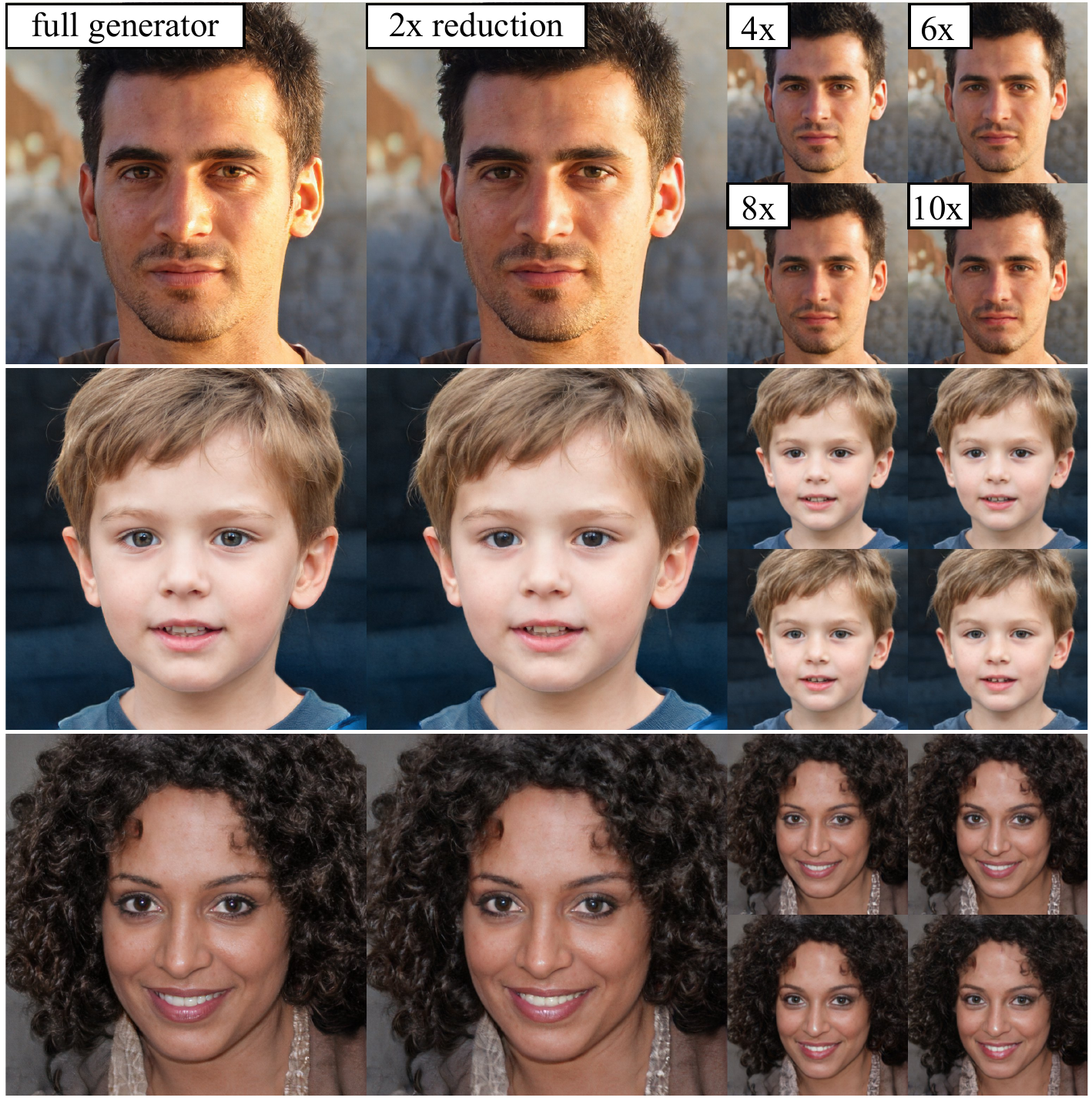}
    \caption{
    \textbf{\method (flexible)} maintains output consistency across \emph{fine-grained} computation reductions.
    Zoom in for better view.  To save space, we display some images in a smaller size. Check Figure~\ref{supp_fig:more_synthesis_ffhq_flex} for a larger view.
    }
    \vspace{-5pt}
    \label{fig:qualititive_flexible}
\end{figure}

In Figure~\ref{fig:qualititive}, we show several samples from our anycost generators (uniform channel setting).
Despite only training a single generator, Anycost GAN maintains output consistency across different resolutions and different channel widths compared to the full generator's output, providing a fast preview when a user is exploring the latent space or applying editing operations.
The samples are generated with truncation rate $\psi=0.5$.

We also provide the results of anycost generator (flexible channels +evolutionary search setting) in Figure~\ref{fig:teaser} and Figure~\ref{fig:qualititive_flexible}. With evolutionary search, we can provide sub-generators at fine-grained computation budgets at 2-10$\times$ reduction. The sub-generators share high consistency for the projected and edited images, paving a way for fast editing preview.

\myparagraph{Quantitative results.}
\begin{table}[t]
    \setlength{\tabcolsep}{3.5pt}
    \caption{Anycost GANs achieves similar or better FIDs/path lengths at various channel widths compared to StyleGAN2, despite training a single, flexible generator. }
    \label{tab:highres_quantitative}
    \centering
    \small{
     \begin{tabular}{clcccccc}
    \toprule
     & & \multicolumn{4}{c}{FID-50k$\downarrow$} & \multicolumn{2}{c}{Path length$\downarrow$} \\
     \cmidrule(lr){3-6} \cmidrule(lr){7-8}
     \multicolumn{2}{c}{Channels:} & 1.0$\times$ & 0.75$\times$ & 0.5$\times$ & 0.25$\times$ & 1.0$\times$ & 0.5$\times$ \\ 
     \midrule
    \multirow{2}{*}{\shortstack{FFHQ\\1024}} & StyleGAN2 & 2.84 & - & 3.31 & - & 145.0 & 124.5\\ 
    \cmidrule(lr){2-8}
    & Anycost & 2.77 & 3.05 & 3.28 & 5.01 & 144.2 & 147.2 \\
    \midrule
    \multirow{2}{*}{\shortstack{Car\\512}} & StyleGAN2 & 2.32 & - & 3.19 & - & 415.5 & 471.2\\
    \cmidrule(lr){2-8}
    & Anycost & 2.38 & 2.46 & 2.61 & 3.69 & 380.1 & 430.0 \\
    \bottomrule
     \end{tabular}
     }
\end{table}
We also provide the quantitative results of the high-resolution images. The high-resolution FIDs and path lengths~\cite{karras2019style} are shown in Table~\ref{tab:highres_quantitative}. We only provide the results of uniform channel setting, since the flexible setting shares a similar FID \vs computation trade-off (Figure~\ref{fig:fid_vs_macs}). Compared to StyleGAN2, Anycost GANs achieves similar or better FIDs and path lengths at various channel widths, despite only one generator is trained. Anycost GANs enjoy better flexibility for inference and also better synthesis quality compared to the baselines.

\begin{figure*}[t!]
    \centering
    \includegraphics[width=0.95\textwidth]{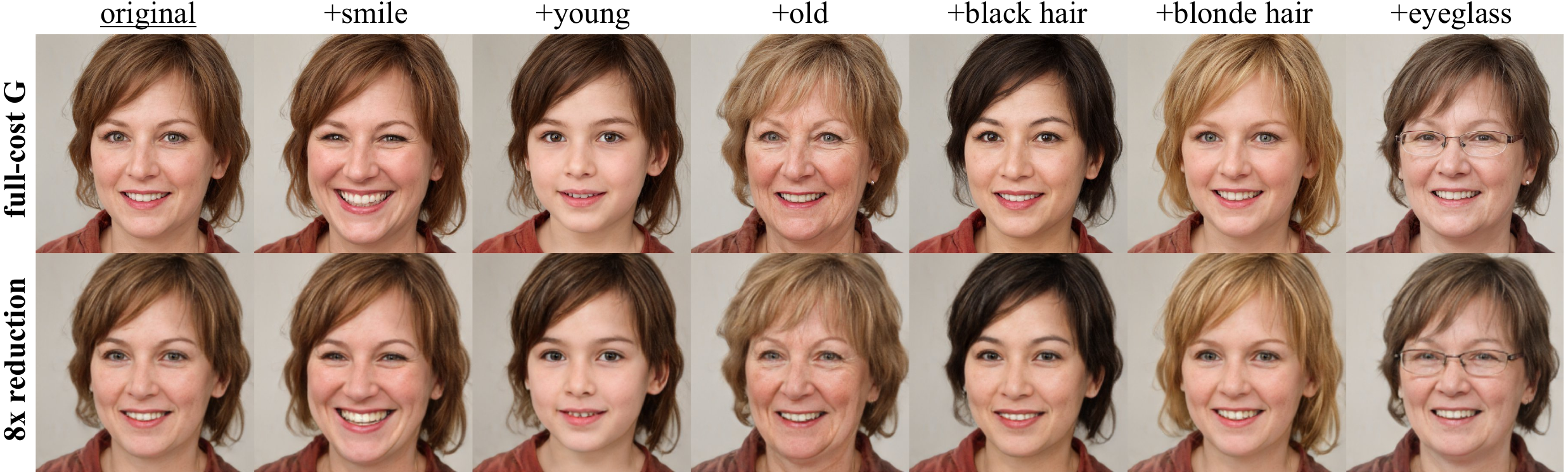}
    \caption{\method maintains consistency after image editing, providing a quick preview at 8$\times$ computation reduction.}
    \label{fig:compare_edited}
\end{figure*}

\myparagraph{Attribute consistency.}

\begin{table}[t]
    \setlength{\tabcolsep}{2pt}
    \caption{The generated outputs of the full and sub-generators have high \emph{attribute consistency}, validated by match rates. The chosen attributes have high inconsistency for two separately trained generators (``Separate'').}
    \label{tab:attribute}
    \centering
    \small{
     \begin{tabular}{rcccccc}
    \toprule
       & Smiling  & BlackHair & Eyeglass & StraightHair & Earrings \\ 
       \midrule
      Separate & 55.8\% & 62.7\% & 83.4\% & 55.8\% & 50.6\% \\
      \midrule
    0.5$\times$ ch & 98.1\%  & 97.0\% & 99.9\% & 95.9\%  & 94.1\% \\ 
    0.25$\times$ ch & 96.9\%   & 95.5\% & 99.8\% & 93.6\% & 88.4\% \\ 
    \bottomrule
     \end{tabular}
     }
\end{table}
Our low-cost results not only preserve the visual consistency (LPIPS difference),  but also depict similar visual attributes as the full results do.
We check the binary attributes (\eg, is smiling, is wearing eyeglasses) of images generated by sub-generators (0.5$\times$ and 0.25$\times$ channels) and the full generator. 
Following~\cite{karras2019style}, we trained the attribute classifier to predict 40 attributes 
on CelebA-HQ for evaluation. We compare the prediction results on 10,000 randomly generated images and report the match rates in Table~\ref{tab:attribute}. For all the categories, we can achieve $>95\%$ consistency with 0.5$\times$ channel  (except for ``Earrings'', which is challenging due to its small size), which is higher than the accuracy of attribute classifiers on CelebA~\cite{han2017heterogeneous}, indicating that our model preserves high-level visual attributes across full generators and sub-generators. The attributes in Table~\ref{tab:attribute} are chosen due to their high inconsistency between two separately trained generators (``Separate'').
\myparagraph{Latency reduction.}

\begin{table}[t]
    \setlength{\tabcolsep}{4pt}
    \caption{Inference FPS \& speed up rates on different devices. }
    \label{tab:latency}
    \centering
    \small{
     \begin{tabular}{lcccccc}
    \toprule
     \textbf{FLOPs reduction} & 1$\times$  & 2$\times$ & 4$\times$ & 6$\times$& 8$\times$ & 10$\times$\\ \midrule
     \textbf{Xeon CPU} FPS & 0.63 & 1.78 & 5.94 & 6.24 & 7.48 & 7.35 \\  
     speed up rate & 1$\times$  & 2.8$\times$  & 9.5$\times$  & 10.0$\times$  & 11.9$\times$  & 11.7$\times$  \\ \midrule
     \textbf{Nano GPU} FPS & 0.65 &	1.17 & 2.77	& 3.59 & 4.07 & 4.1 \\
     speed up rate & 1$\times$ &  1.8$\times$ & 4.2$\times$ & 5.5$\times$ & 6.2$\times$ &	6.3$\times$ \\
    \bottomrule
     \end{tabular}
     }
\end{table}

We measure the latency speed-up on both desktop CPU (Intel Xeon Silver) and mobile GPU (NVIDIA Jetson Nano) in Table~\ref{tab:latency}.
Anycost generators can generate consistent previews at 11.9$\times$ walltime speed-up on Xeon CPU and 8.5$\times$ speed-up on Jetson Nano. Interestingly, the model achieves a super-linear speed-up on Intel CPU. We hypothesize that the activation and weight sizes of the full-cost generator exceed the cache on CPU (16.5MB), leading to increased cache miss rate and worse inference efficiency.

\subsection{Anycost Image Projection and Editing}
\vspace{4pt}
\myparagraph{Encoder training.}
\begin{table}[t]
    \setlength{\tabcolsep}{3pt}
    \caption{Consistency-aware encoder gives accurate projection results for both the full generator and sub-generators. It largely reduces the gap between them. Here $\mathbf{w}=E(x)$. }
    \label{tab:encoder}
    \centering
    \small{
     \begin{tabular}{lcccccccc}
    \toprule
       & \multicolumn{2}{c}{$\ell(G(\mathbf{w}), \image)$} & \multicolumn{2}{c}{$\ell(G'(\mathbf{w}), \image)$} & \multicolumn{2}{c}{$\ell(G'(\mathbf{w}), G(\mathbf{w}))$}  \\  
       \cmidrule(lr){2-3} \cmidrule(lr){4-5} \cmidrule(lr){6-7}
       & LPIPS & MSE & LPIPS & MSE & LPIPS & MSE\\
      \midrule
      ALAE~\cite{pidhorskyi2020adversarial} & 0.32 & 0.15 & - & - & - & -\\
      IDInvert~\cite{zhu2020indomain} & 0.22 & 0.06 & - & - & - & -\\
      pSp~\cite{richardson2020encoding} & 0.19 & 0.04 & - & - & - & -\\
      \midrule
      Ours (fullG-only) & \textbf{0.13} & \textbf{0.03} & 0.17 & 0.04 & 0.07 & 0.012 \\
      Ours (full+subG) & \textbf{0.13} & \textbf{0.03} & \textbf{0.14} & \textbf{0.03} & \textbf{0.02} & \textbf{0.003}\\
    \bottomrule
     \end{tabular}
     }
\end{table}
We compare the encoder only optimized for the full model and the one optimized for both the full network and the sub-networks. The results on CelebA test set are in Table~\ref{tab:encoder}. For generality, we used a simple encoder architecture: we used the ResNet-50~\cite{he2016deep} backbone architecture and a linear layer to regress the latent code in $\mathcal{W}+$ space. We train the encoder both on real images (FFHQ + CelebA train set following~\cite{richardson2020encoding}) and generated images for 200 epochs. We apply random horizontal flip, random color jittering, and random grayscale as augmentation~\cite{he2020momentum}. To compare with existing literature, we measure the LPIPS loss using AlexNet backbone instead of VGG. Apart from the reconstruction loss for the full generator, we also measure the average reconstruction performance for all the sub-generator architectures found by our evolutionary algorithm. 
Our generator has better reconstruction performance compared to an advanced encoder design~\cite{richardson2020encoding} that uses a Feature Pyramid Network~\cite{lin2017feature} structure.

\myparagraph{Optimization-based projection.}
For optimization-based projection, we used L-BFGS solver~\cite{liu1989limited}  with 100 iterations. We  find that L-BFGS converges faster than Adam~\cite{kingma2014adam}. We use the encoder's prediction as the starting point to represent a more realistic use case, which also results in better convergence compared to starting from average latent $\mathbf{w}$~\cite{karras2020analyzing}. Interestingly, we find that a lower optimization loss results in a better reconstruction, but the latent code may not be suitable for latent space-based editing (see Section~\ref{sec:supp_optimization_vs_editing}). Therefore, we do not benchmark the quantitative results here. The qualitative results are shown in Figure~\ref{fig:teaser}.

\myparagraph{Image editing with anycost generators.}

We show that anycost generators remain consistent after image editing. We compute several editing directions on FFHQ dataset in the $\mathcal{W}$ space following~\cite{shen2020interpreting}. We compare the outputs of the full generators and 8$\times$-smaller computation sub-generators after editing smiling, age, eye-glasses, and hair color in the latent space. The results are shown in \reffig{compare_edited} and \reffig{teaser} . Anycost generator gives consistent outputs under various attribute editing.
We show more projection and editing examples in the supplementary materials.

\section{Discussion}
In this paper, we have proposed \method, a scalable training method for learning unconditional generators that can adapt to diverse hardware and user latency requirements. We have demonstrated its application in image projection and editing. Several limitations still exist for our method. First, the control over the channel numbers might be difficult for users who are new to neural networks. In the future, we plan to provide more intuitive controls such as synthesizing similar color, texture, illumination, or shape as the original model does. Second, our current model aims to approximate every single output pixel equally well. In practice, certain objects (e.g., face) might be more important than others (e.g., background). We would like to learn models that can support spatially-varying trade-off between fidelity and latency.%

\myparagraph{Acknowledgment. } 
We thank Taesung Park and Zhixin Shu for the helpful discussion.
Part of the work is supported under NSF CAREER Award \#1943349. We thank MIT-IBM Watson AI Lab for the support.

\clearpage
{\small
\bibliographystyle{ieee_fullname}
\bibliography{main}
}

\clearpage
\appendix

\section{Interactive Demo}

\begin{figure}[h]
    \centering
    \includegraphics[width=0.5\textwidth]{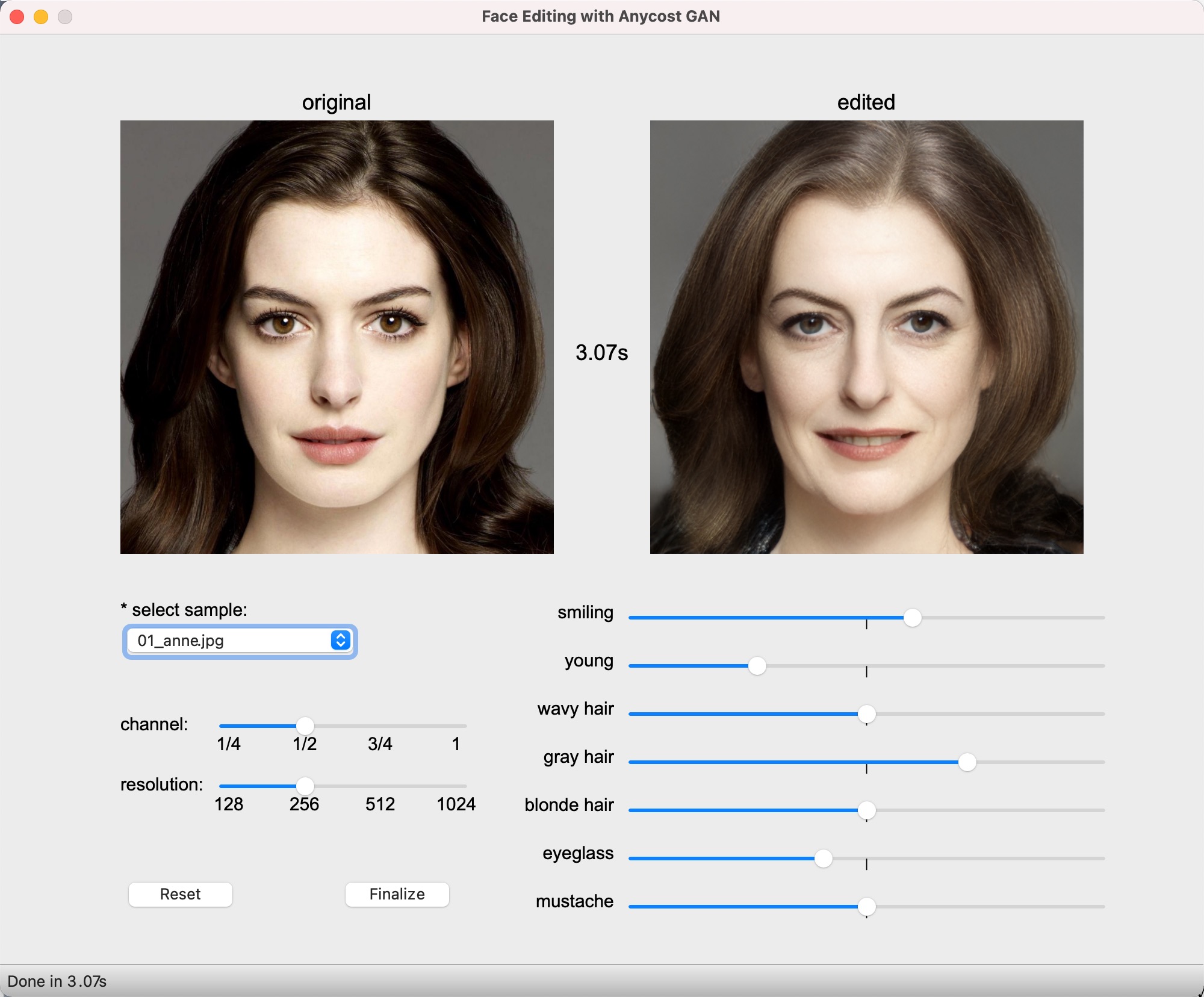}
    \vspace{-8pt}
    \caption{The UI interface of our face editing demo. Please refer to our demo video for more details. }
    \vspace{-5pt}
    \label{supp_fig:demo_interface}
\end{figure}

We provide an interactive demo in the code release to show the image editing use case with our anycost generator. Anycost generator provides a fast preview to enable interactive image editing. A screenshot of the demo interface is provided in Figure~\ref{supp_fig:demo_interface}.

\section{Experimental Details}
\label{sec: supp_exp_details}

\subsection{Training}
\label{sec: supp_training}

We used similar training hyper-parameters as~\cite{karras2020analyzing}. For the first stage, \ie, training a vanilla StyleGAN2 model, we used exactly the same training configuration. For multi-resolution training and adaptive-channel training, since we can distill from the pre-trained generator, we do not need to add the path length regularization~\cite{karras2020analyzing}, which accelerates the training. 

For multi-resolution training, at reach iteration, we randomly sampled two resolutions from the four possible resolutions for both generator and discriminator training, which improves the data efficiency and stabilizes training. During adaptive-channel training, we randomly sampled one channel configuration at each iteration. For the uniform channel setting, we randomly sampled one of the four ratios using a uniform probability. For the flexible channel setting, we followed~\cite{li2017universal} to use a ``sandwich'' rule for channel sampling (\ie, sample the full generator for 25\% of the time, sample the smallest generator for 25\% of the time, and randomly sample a generator at the rest of the time). We empirically find that it makes the convergence more stable and improves the FID of the full generator.

\subsection{Evolutionary search}
\label{sec: supp_evolve}
We use evolutionary search to find the best sub-generator under a certain resource budget. We use a population size $P=50$ and max iterations $\mathcal{T}=20$. For each iteration, we only keep the top-10 best candidates in the current generation, and then generate a new generation consisting of 25 \emph{crossover} candidates and 25 \emph{mutation} candidates. %
For a crossover, we randomly sample two top candidates and cross them to generate a new one; for mutation, we keep a mutation probability of 0.1. For the new candidates, 
We directly evaluate the output difference \wrt the full generator or FIDs without further fine-tuning, as our anycost generator is well trained for any subset of weights.%

\subsection{Metrics}

\vspace{5pt}
\myparagraph{Fr\'echet Inception Distance (FID). }
For FID~\cite{heusel2017gans} evaluation on high-resolution images (1024 for FFHQ~\cite{karras2020analyzing} and 512 for LSUN Car~\cite{yu2015lsun}), we follow~\cite{karras2020analyzing} to compute the FID using 50k images (\ie, FID-50k). For FID evaluation on low-resolution images ($\leq$ 256 on FFHQ), we follow~\cite{karras2020training} to compute the FID using 70k images (\ie, FID-70k), so that we can compare with the results in the paper.

\myparagraph{Perceptual path length (PPL).} We follow~\cite{karras2020analyzing} to compute the PPL metric, using 100k random samples.

\myparagraph{Attribute consistency.} We trained our own attribute predictor to evaluate the attribute consistency (can be found in the code release). We randomly sampled 10k images with truncation $\psi=0.5$ to compute the attribute consistency.

\section{Training Time}

We discuss the training time on FFHQ~\cite{karras2020analyzing} dataset that consists of 70k samples. Our experiments are based on a PyTorch implementation of StyleGAN2~\cite{karras2020analyzing}\footnote{https://github.com/rosinality/stylegan2-pytorch}. Given a normally trained StyleGAN2 model, we first train the model for 5 million images to support multi-resolution, and then train it for another 5 million images to support adaptive channel numbers. On 8 NVIDIA Titan RTX GPUs, the training takes 5.3 
days in total, which is roughly 60\% of the StyleGAN2 training time.
Note that we just need to train the model once, and we can run it under various computational budgets by using a smaller sub-generator, while distillation or compression-based methods need to train each sub-generator architecture one by one.

\begin{figure}[h]
    \centering
    \includegraphics[width=0.32\textwidth]{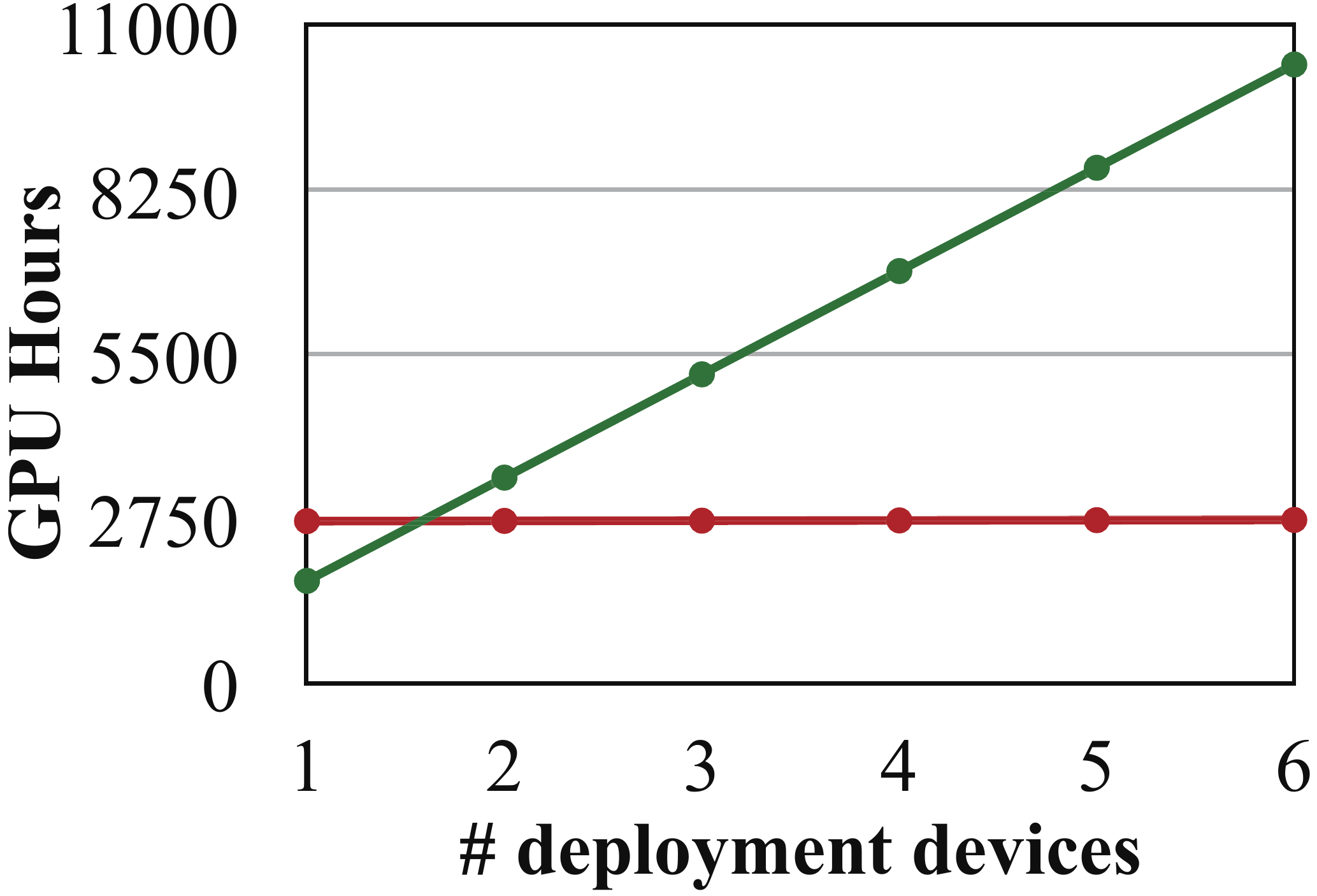}
    \vspace{-8pt}
    \caption{Anycost GAN has a small marginal cost for supporting extra devices. }
    \vspace{-5pt}
    \label{supp_fig:gpu_hours}
\end{figure}

The cost of evolutionary search is small compared to the training. We used 2048 samples to evaluate the difference between the full generator and sub-generators' outputs and find the one with minimal difference. Given a certain budget, the evolutionary search can be done in 4 hours on a single GPU. Therefore, the marginal cost to support extra devices is very small. We compare the GPU hours \vs \#devices in Figure~\ref{supp_fig:gpu_hours}. We can save $3.8\times$ GPU hours when deploying on 6 different devices. This is a practical case when we cover devices from smartphones, iPads, laptops, to cloud servers.

\section{Image Projection and Editing}

\subsection{Extracting Editing Directions}
\label{sec:extract_directions}
To extract the editing directions, We use the trained generator to randomly generate 100,000 $\{\mathbf{w}, G(\mathbf{w})\}$ pairs, and use a pre-trained attribute classifier to give the binary attribute predictions (\eg, young or old)~\cite{shen2020interpreting}. The samples are generated using truncation $\psi=0.7$.
We then compute the decision boundary between positive $\mathbf{w}$'s and negative $\mathbf{w}$'s.
Please refer to~\cite{shen2020interpreting} for more details.
We empirically find that, finding the editing directions is less precise compared to image projection. Therefore, we only used the largest generator to compute the directions, which works pretty well for other sub-generators.

\subsection{Discussion on Projection \vs Editing}
\label{sec:supp_optimization_vs_editing}

\begin{figure}[t]
    \centering
    \includegraphics[width=0.45\textwidth]{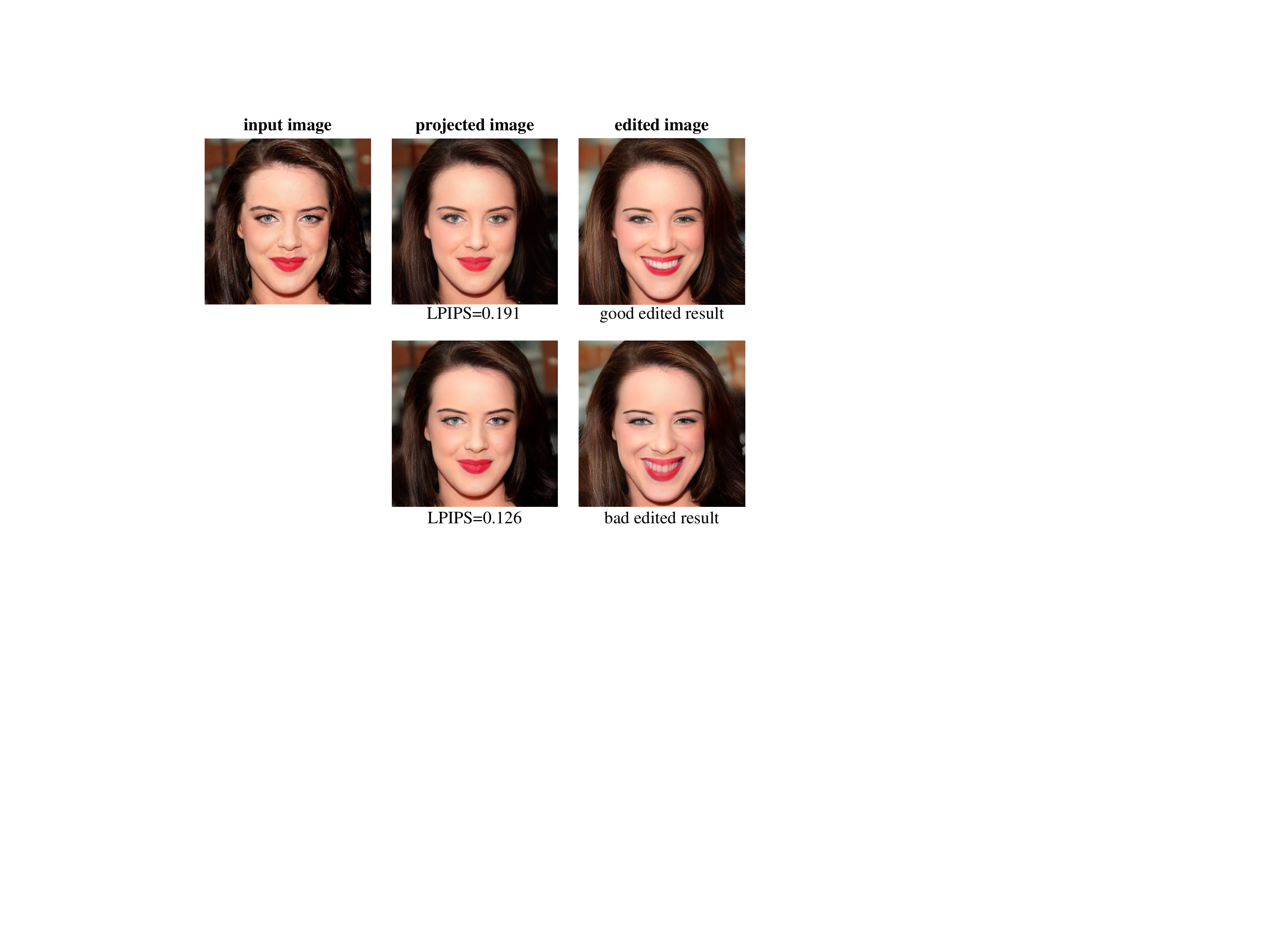}
    \vspace{-8pt}
    \caption{A better projected latent code (\ie, with lower LPIPS loss) is not necessarily easily editable. The second projection (row 2) has lower LPIPS loss, but after editing, there are clear artifacts. }
    \vspace{-5pt}
    \label{supp_fig:project_vs_edit}
\end{figure}

We notice that sometimes a better projected image may not lead to an easily editable latent code. For example, in Figure~\ref{supp_fig:project_vs_edit}, we compare two projected latent codes (row 1 and row 2). The second projected code has lower LPIPS loss, but after editing, there are clear artifacts on the image. We hypothesize that the projected latent code may not be in the training domain of the generator~\cite{zhu2020indomain}. Therefore, the reconstruction loss alone cannot fully reflect the quality of the projected code in image editing scenario. We left the more in-depth discussion to the future work.

In practice, we empirically find that encoder-based initialization and 100 iterations of backward optimization can lead to both good visual similarity and editing ability at the same time.

\subsection{Qualitative Comparison of Image Projection }

We have shown the quantitative results for image projection in the main paper (Table~\ref{tab:encoder}). Here we compare the qualitative results of baseline image projection and consistency-aware projection to show the necessity of applying consistency constraints.

\paragraph{Encoder-based projection.}

\begin{figure}[t]
    \centering
    \includegraphics[width=0.5\textwidth]{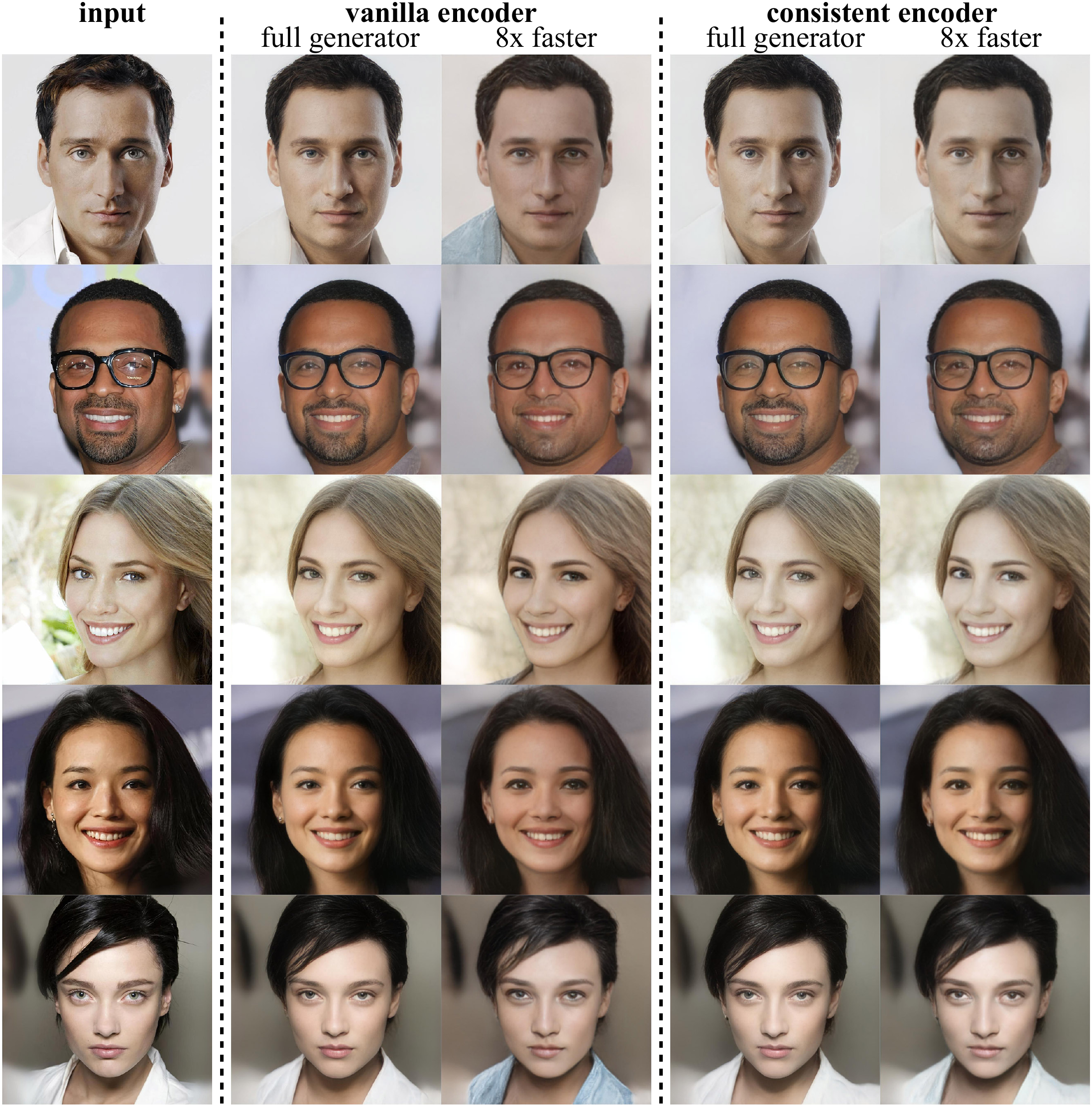}
    \vspace{-8pt}
    \caption{Our consistent-aware encoder achieves better consistency between the full and the sub-generator's outputs. Vanilla encoder trained only for the full generator leads to visible differences in clothing colors, skin tones, and more fine-grained details such as  eye shapes. Both encoders have achieved the same reconstruction performance for the full generator.Results are hand-picked to reflect the difference. }
    \vspace{-5pt}
    \label{supp_fig:compare_encoder}
\end{figure}

We compare the baseline encoder optimized only for the full generator, and our consistency-aware encoder optimized for both full generator and sub-generators. Both encoders achieve the same level of reconstruction performance for the full generator (measured by LPIPS+MSE loss). 
Given the encoder's predicted $\mathbf{w}$ code, we compare the outputs of the full generator and the 8$\times$ faster sub-generator.
The results are shown in Figure~\ref{supp_fig:compare_encoder}. We can see that our consistency-aware encoder results in a smaller gap between the full generator and sub-generators' outputs. Vanilla encoder trained only for the full generator leads to a poor reconstruction performance for the 8$\times$ faster generator, and a noticeable difference between the two outputs: there are differences in both macro properties like skin tones, clothing colors, and fine-grained details such as eye shapes. The comparison demonstrates the advantage for our consistency-aware encoder training. \emph{Note that the images are picked to clearly demonstrate the difference.}

\paragraph{Optimization-based projection.} 

\begin{figure*}[t]
    \centering
    \includegraphics[width=0.85\textwidth]{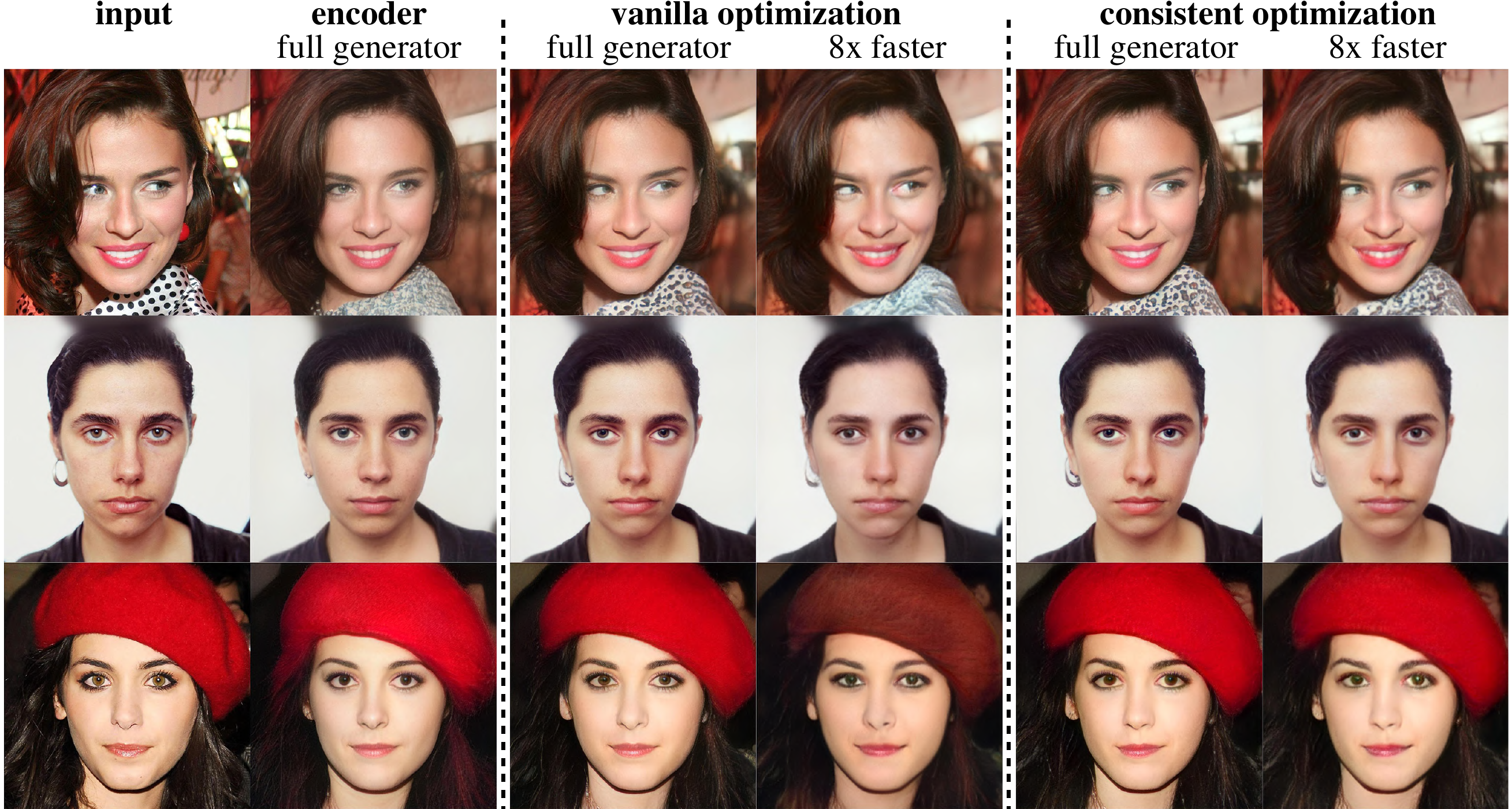}
    \vspace{-8pt}
    \caption{Consistent-aware optimization leads to better image projection for sub-generators, reducing the output difference between the full generator and sub-generators. Results are hand-picked to reflect the difference. }
    \vspace{-5pt}
    \label{supp_fig:compare_optimization}
\end{figure*}

Given the encoder's prediction, we further optimize $\mathbf{w}$ to improve the projection quality. 
We provide a visual comparison of vanilla optimization and consistency-aware optimization in Figure~\ref{supp_fig:compare_optimization}. Again, we compare the outputs of the full generator and 8$\times$ faster sub-generator.
First, despite that the encoder's predicted $\mathbf{w}$ can reconstruct the input image reasonably well (the second column), additional optimization can help refine details (\eg, the gaze in the first example, the earring in the second example), which is critical for editing high-resolution real images. Second, the consistent-aware optimization provides better consistency between the full and sub-generators' outputs (\eg, the clothing in the first example, and the eye shape in the second examples, the skin tone and cap color in the third example). The improved consistency allows us to provide a more accurate, fast preview for the final output. \emph{Note that the images are picked to clearly demonstrate the difference}.

\section{Additional Synthesis Results}

\vspace{10pt}
\myparagraph{Anycost GAN (uniform) on FFHQ. }

\begin{figure*}[t]
    \centering
    \vspace{-20pt}
    \includegraphics[width=0.9\textwidth]{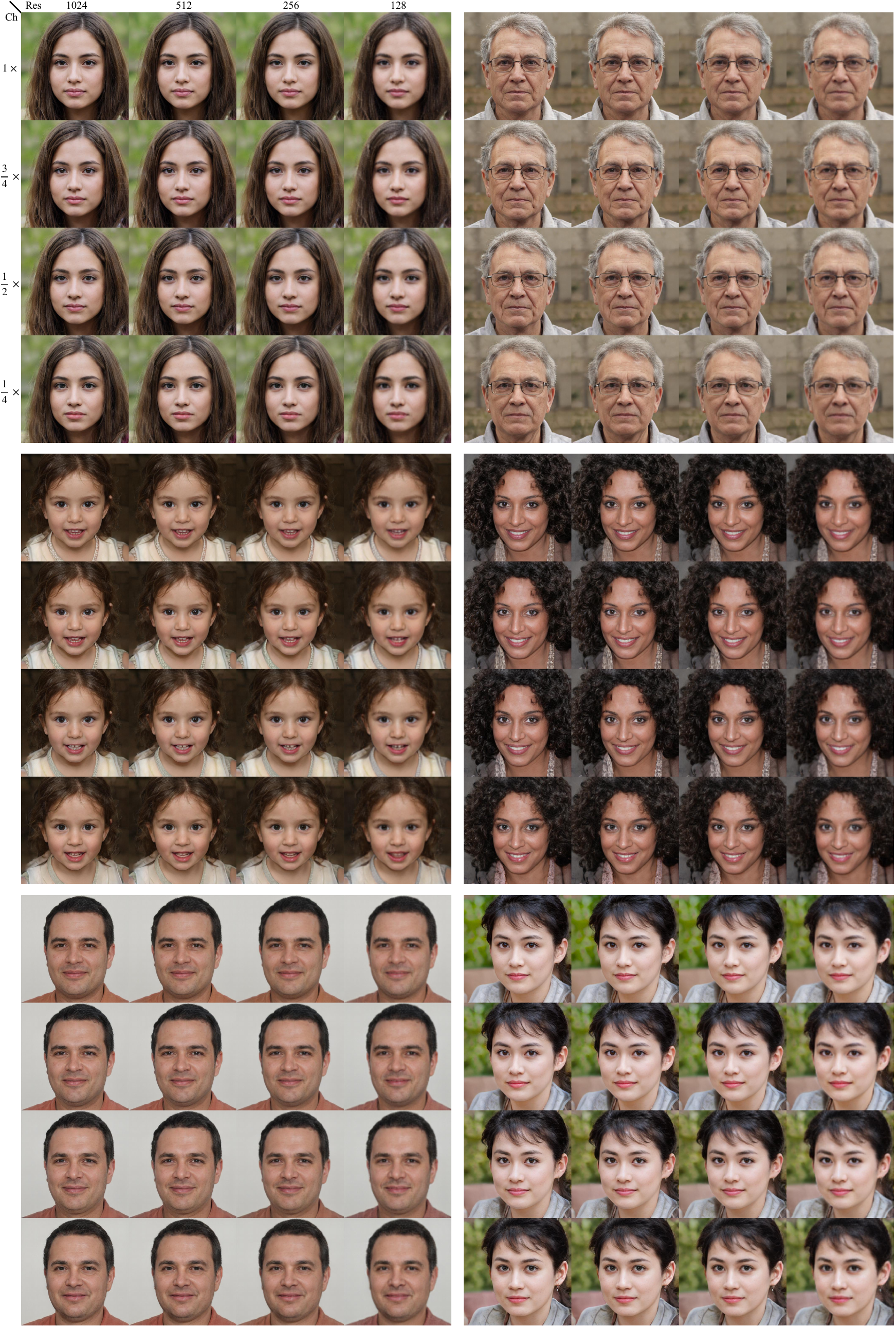}
    \vspace{-8pt}
    \caption{Additional synthesis results for anycost generator (uniform) on FFHQ.}
    \vspace{-5pt}
    \label{supp_fig:more_synthesis_ffhq_uni}
\end{figure*}
We provide more results of Anycost GAN (uniform) on FFHQ dataset~\cite{karras2020analyzing} in Figure~\ref{supp_fig:more_synthesis_ffhq_uni}.

\myparagraph{Anycost GAN (flexible) on FFHQ. }

\begin{figure*}[t]
    \centering
    \vspace{-10pt}
    \includegraphics[width=0.9\textwidth]{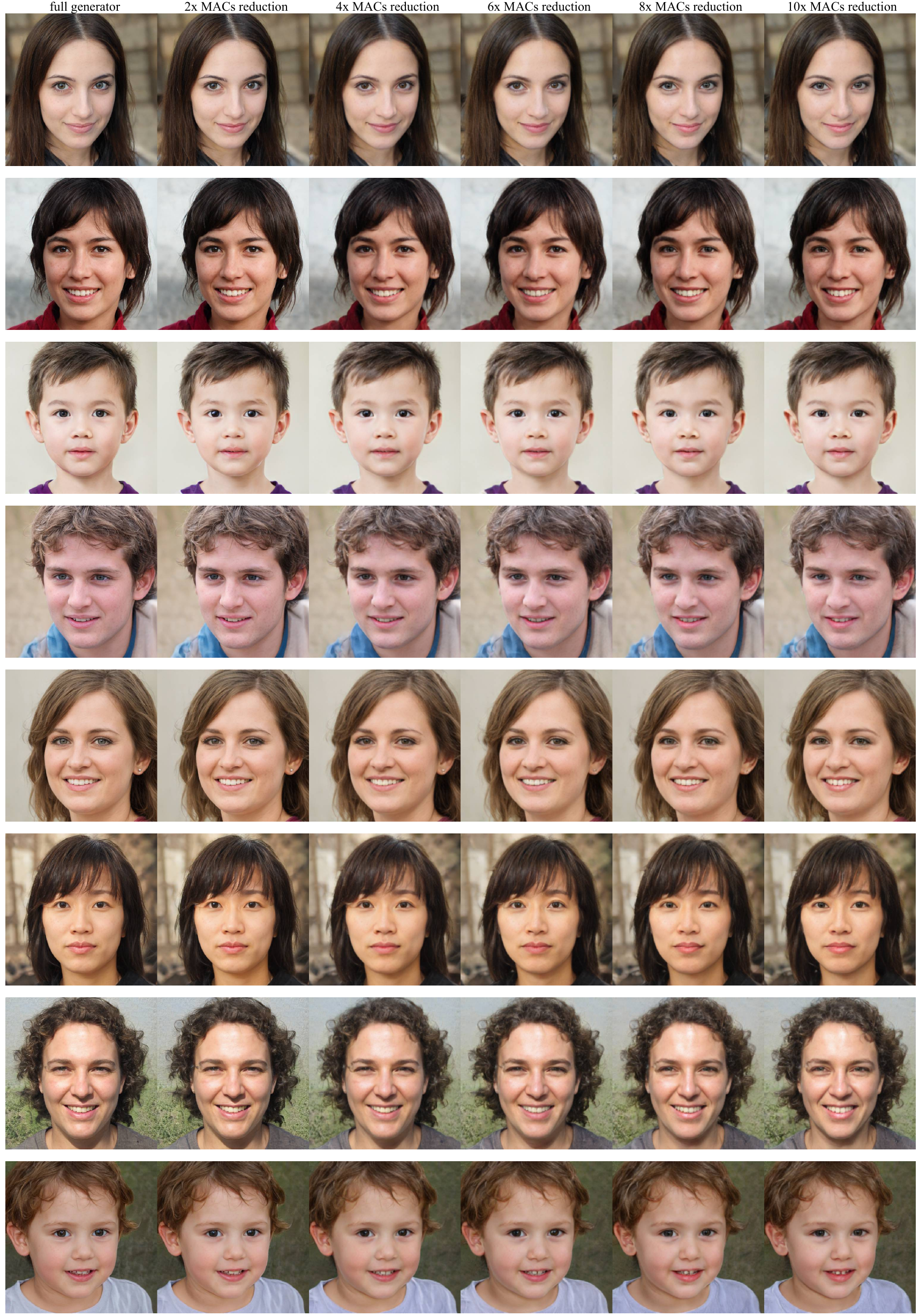}
    \vspace{-8pt}
    \caption{Additional synthesis results for anycost generator (flexible) on FFHQ.}
    \vspace{-5pt}
    \label{supp_fig:more_synthesis_ffhq_flex}
\end{figure*}
We provide more results of Anycost GAN (uniform) on FFHQ dataset~\cite{karras2020analyzing} in Figure~\ref{supp_fig:more_synthesis_ffhq_flex}.

\myparagraph{Anycost GAN (uniform) on LSUN Car.}

\begin{figure*}[t]
    \centering
    \vspace{-20pt}
    \includegraphics[width=0.9\textwidth]{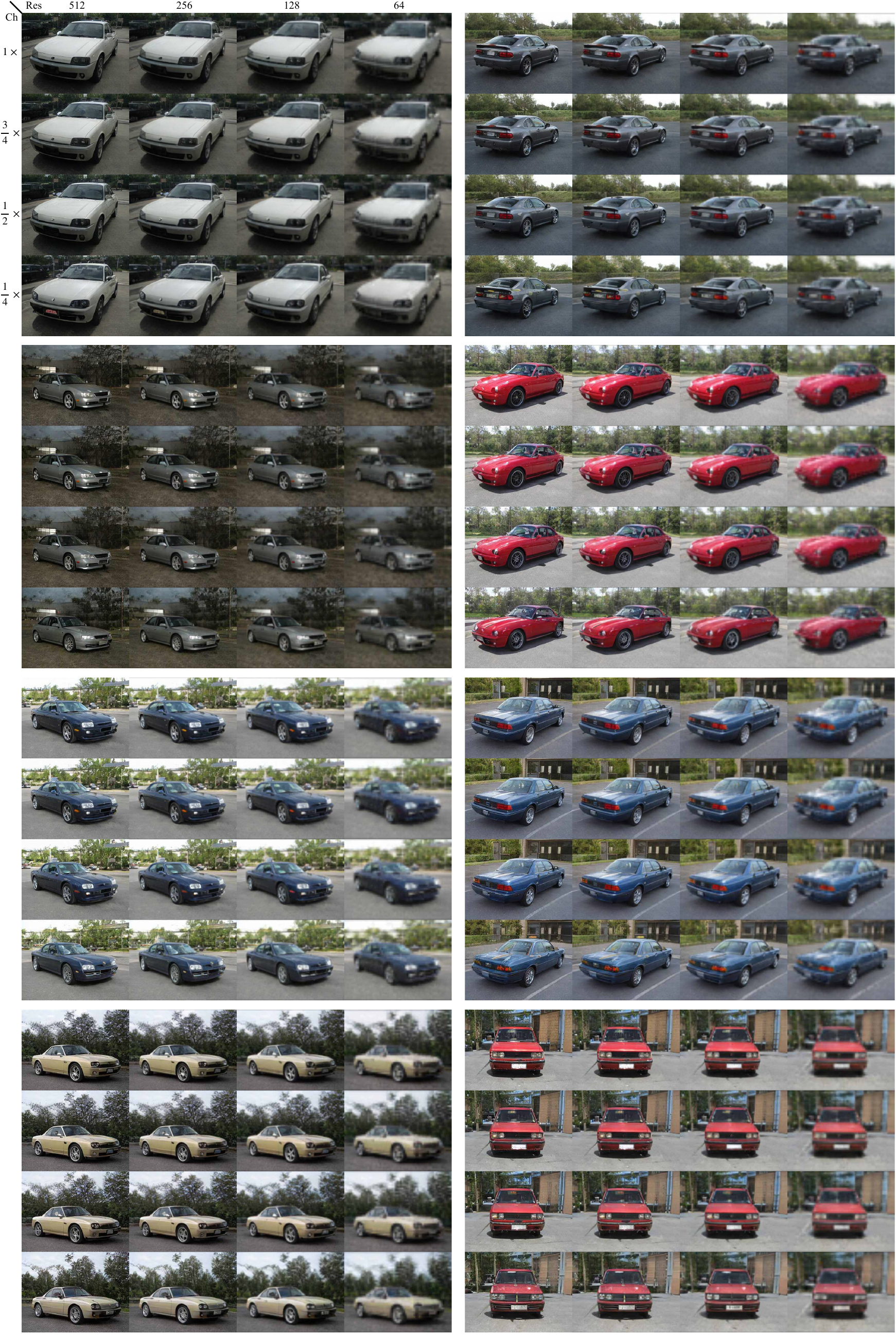}
    \vspace{-8pt}
    \caption{Additional synthesis results for anycost generator (uniform) on LSUN Car.}
    \vspace{-5pt}
    \label{supp_fig:more_synthesis_car_uni}
\end{figure*}
We provide more results of Anycost GAN (uniform) on LSUN Car dataset~\cite{yu2015lsun} in Figure~\ref{supp_fig:more_synthesis_car_uni}.

\section{Additional Projection Results}

\begin{figure*}[t]
    \centering
    \includegraphics[width=0.98\textwidth]{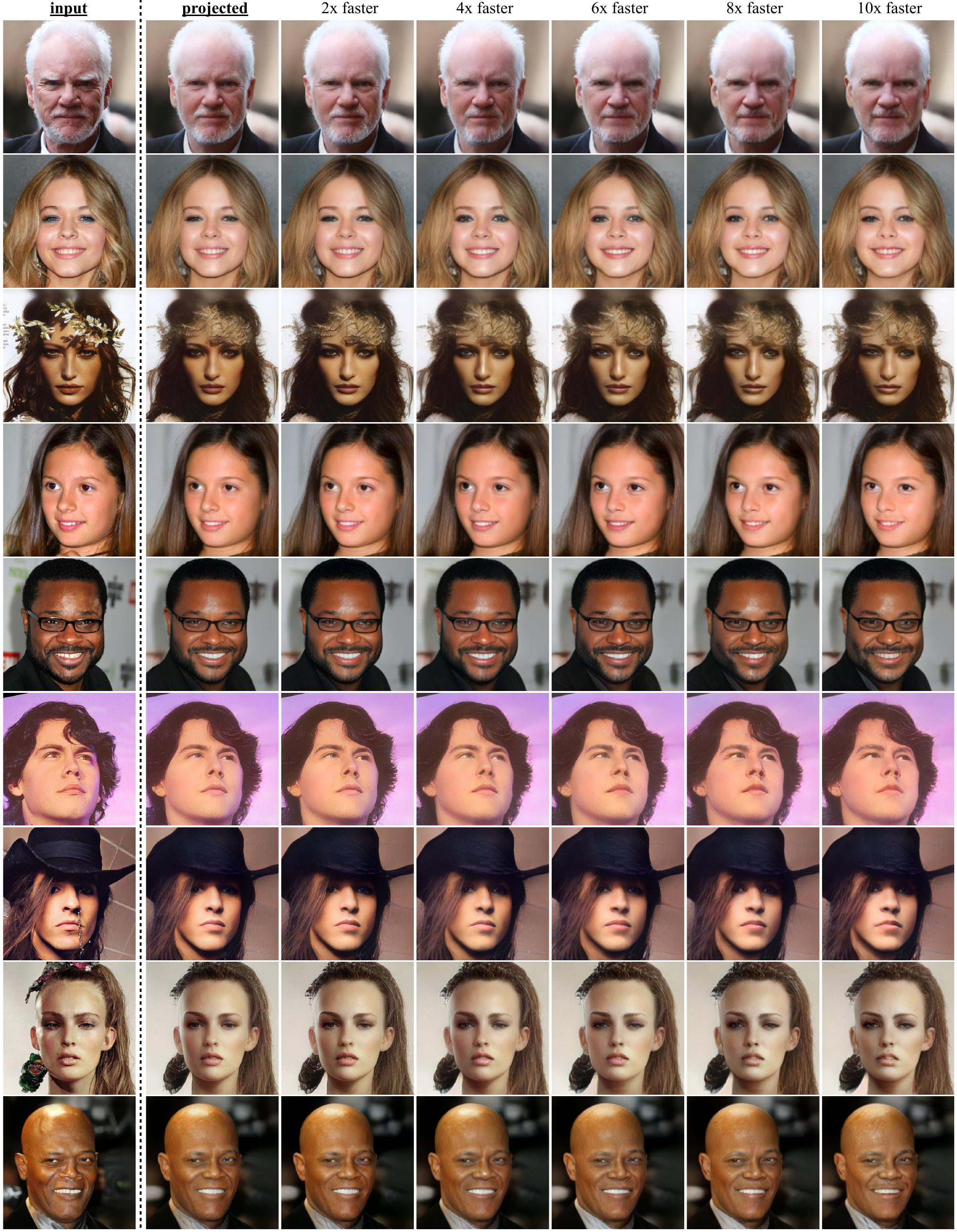}
    \vspace{-8pt}
    \caption{Additional projection results for FFHQ anycost generators. Zoom in for a better view.}
    \vspace{-5pt}
    \label{supp_fig:more_projection_ffhq}
\end{figure*}

We provide additional projection results on FFHQ in Figure~\ref{supp_fig:more_projection_ffhq}. The target images are taken from the test set of CelebA-HQ dataset~\cite{karras2018progressive} dataset. Anycost generators produce consistent outputs at various computational budgets.

\section{Additional Editing Results}

\begin{figure*}[t]
    \centering
    \includegraphics[width=0.98\textwidth]{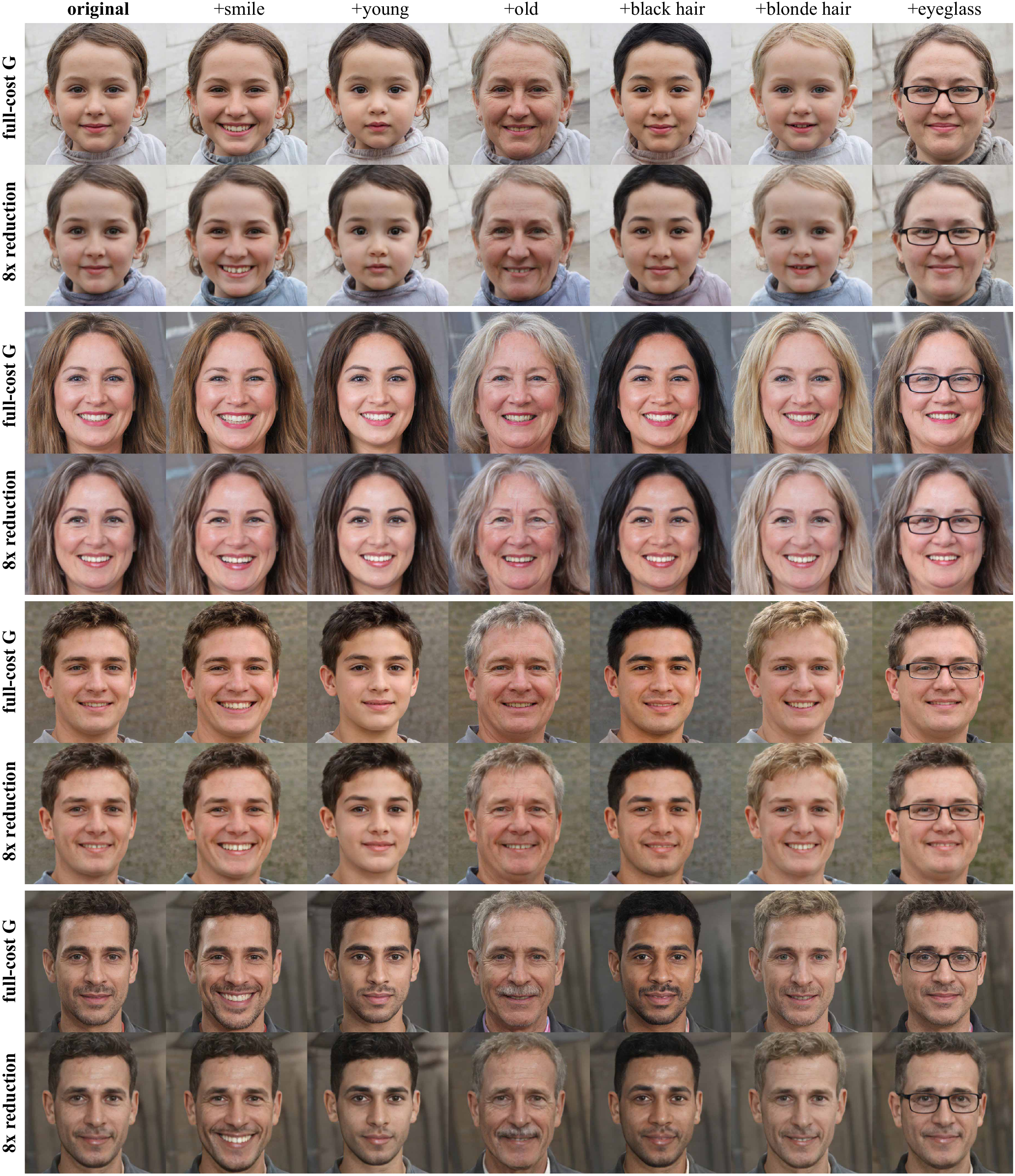}
    \vspace{-8pt}
    \caption{Editing results for FFHQ anycost generator.}
    \vspace{-5pt}
    \label{supp_fig:more_edit_ffhq}
\end{figure*}

We provide more editing results of FFHQ generator. Similarly, we compare the edited images from the full generator and the $8\times$ computation-reduced sub-generator. The results are shown in Figure~\ref{supp_fig:more_edit_ffhq}.

\section{Failure Cases}

\begin{figure}[H]
    \centering
    \includegraphics[width=0.45\textwidth]{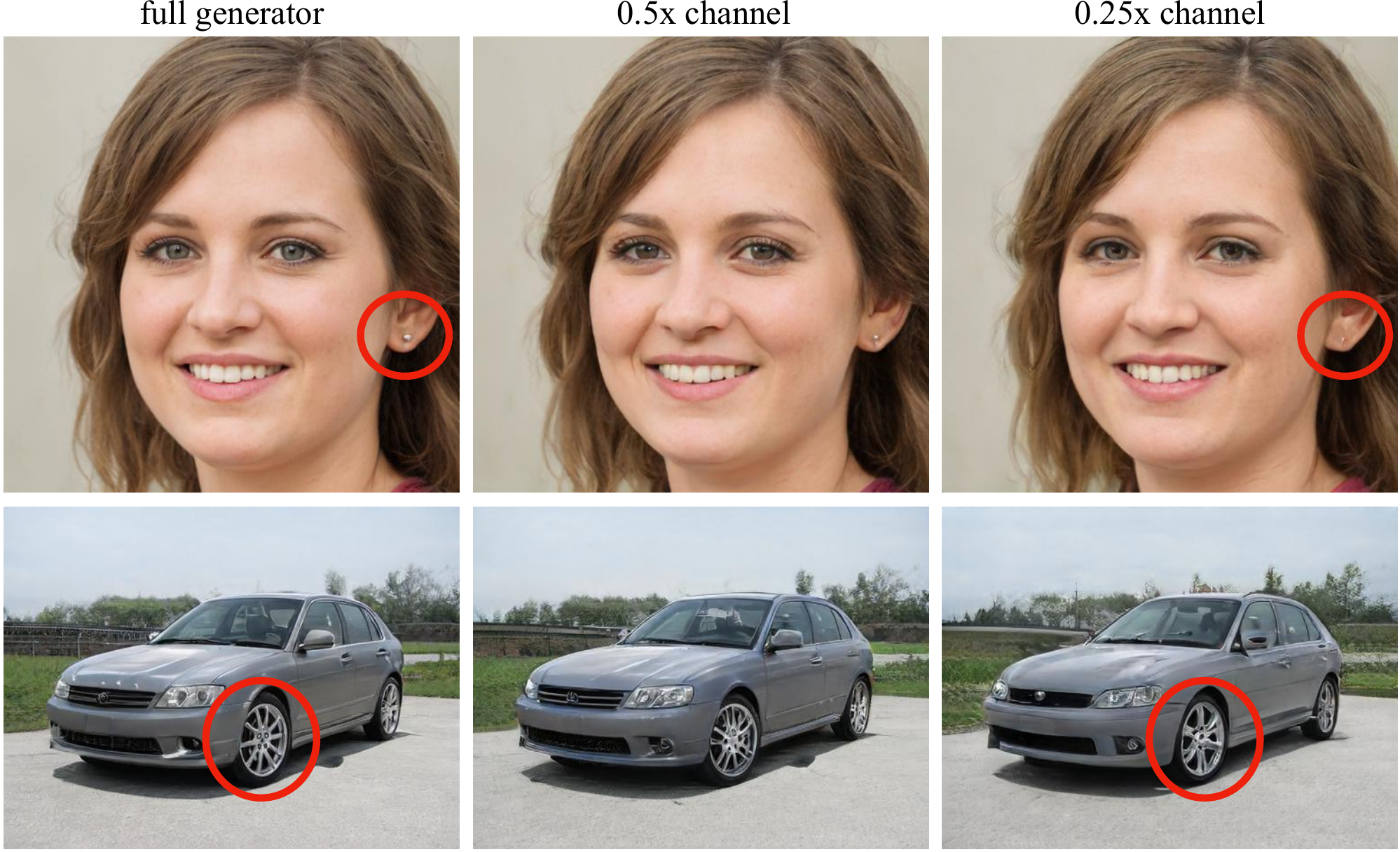}
    \vspace{-8pt}
    \caption{Some fine details are changed when using a sub-generator for synthesis.}
    \vspace{-5pt}
    \label{supp_fig:failure_case}
\end{figure}

The major failure cases we observed are that, some fine details are missing or changed in smaller generators' outputs, like earrings and rim. We show some example images in Figure~\ref{supp_fig:failure_case}. It could be a problem when we try to edit the details of images. The failure case could be mitigated by using a slightly larger sub-generator to provide better consistency (\eg, 0.5$\times$ instead of 0.25$\times$ in this example).

\end{document}